\documentclass[10pt,journal,compsoc]{IEEEtran}
\usepackage{booktabs}
\usepackage{algorithm}
\usepackage{amssymb}
\usepackage{amsfonts}
\usepackage{multirow}
\usepackage{amsmath}
\usepackage{cite}
\usepackage[dvips]{graphicx}
\usepackage{color}
\usepackage{comment}
\usepackage{makecell}
\usepackage{subfigure}
\usepackage{thmtools, thm-restate}
\usepackage{graphicx}
\usepackage{caption}

\makeatletter
\newcommand*\bigcdot{\mathpalette\bigcdot@{.5}}
\newcommand*\bigcdot@[2]{\mathbin{\vcenter{\hbox{\scalebox{#2}{$\m@th#1\bullet$}}}}}
\makeatother

\begin{document}
\newtheorem{definition}{\it Definition}
\newtheorem{theorem}{\bf Theorem}
\newtheorem{lemma}{\it Lemma}
\newtheorem{corollary}{\it Corollary}
\newtheorem{remark}{\it Remark}
\newtheorem{example}{\it Example}
\newtheorem{case}{\bf Case Study}
\newtheorem{assumption}{\it Assumption}
\newtheorem{property}{\it Property}
\newtheorem{proposition}{\it Proposition}

\newcommand{\hP}[1]{{\boldsymbol h}_{{#1}{\bullet}}}
\newcommand{\hS}[1]{{\boldsymbol h}_{{\bullet}{#1}}}

\newcommand{\ba}{\boldsymbol{a}}
\newcommand{\baq}{\overline{q}}
\newcommand{\bA}{\boldsymbol{A}}
\newcommand{\bb}{\boldsymbol{b}}
\newcommand{\bB}{\boldsymbol{B}}
\newcommand{\bc}{\boldsymbol{c}}
\newcommand{\bcO}{\boldsymbol{\cal O}}
\newcommand{\bh}{\boldsymbol{h}}
\newcommand{\bk}{\boldsymbol{k}}
\newcommand{\bH}{\boldsymbol{H}}
\newcommand{\bl}{\boldsymbol{l}}
\newcommand{\bm}{\boldsymbol{m}}
\newcommand{\bn}{\boldsymbol{n}}
\newcommand{\bo}{\boldsymbol{o}}
\newcommand{\bO}{\boldsymbol{O}}
\newcommand{\bp}{\boldsymbol{p}}
\newcommand{\bq}{\boldsymbol{q}}
\newcommand{\bR}{\boldsymbol{R}}
\newcommand{\bs}{\boldsymbol{s}}
\newcommand{\bS}{\boldsymbol{S}}
\newcommand{\bT}{\boldsymbol{T}}
\newcommand{\bw}{\boldsymbol{w}}
\newcommand{\bz}{\boldsymbol{z}}

\newcommand{\balpha}{\boldsymbol{\alpha}}
\newcommand{\bbeta}{\boldsymbol{\beta}}
\newcommand{\bOmega}{\boldsymbol{\Omega}}
\newcommand{\bTheta}{\boldsymbol{\Theta}}
\newcommand{\bphi}{\boldsymbol{\phi}}
\newcommand{\btheta}{\boldsymbol{\theta}}
\newcommand{\bvarpi}{\boldsymbol{\varpi}}
\newcommand{\bpi}{\boldsymbol{\pi}}
\newcommand{\bpsi}{\boldsymbol{\psi}}
\newcommand{\bxi}{\boldsymbol{\xi}}
\newcommand{\bx}{\boldsymbol{x}}
\newcommand{\by}{\boldsymbol{y}}

\newcommand{\cA}{{\cal A}}
\newcommand{\bcA}{\boldsymbol{\cal A}}
\newcommand{\cB}{{\cal B}}
\newcommand{\cE}{{\cal E}}
\newcommand{\cG}{{\cal G}}
\newcommand{\cH}{{\cal H}}
\newcommand{\bcH}{\boldsymbol {\cal H}}
\newcommand{\cK}{{\cal K}}
\newcommand{\cO}{{\cal O}}
\newcommand{\cR}{{\cal R}}
\newcommand{\cS}{{\cal S}}
\newcommand{\dcS}{\ddot{{\cal S}}}
\newcommand{\ds}{\ddot{{s}}}
\newcommand{\cT}{{\cal T}}
\newcommand{\cU}{{\cal U}}
\newcommand{\wt}[1]{\widetilde{#1}}

\newcommand{\mA}{\mathbb{A}}
\newcommand{\mE}{\mathbb{E}}
\newcommand{\mG}{\mathbb{G}}
\newcommand{\mR}{\mathbb{R}}
\newcommand{\mS}{\mathbb{S}}
\newcommand{\mU}{\mathbb{U}}
\newcommand{\mV}{\mathbb{V}}
\newcommand{\mW}{\mathbb{W}}

\newcommand{\uq}{\underline{q}}
\newcommand{\ubq}{\underline{\boldsymbol q}}

\newcommand{\red}[1]{\textcolor[rgb]{1,0,0}{#1}}
\newcommand{\gre}[1]{\textcolor[rgb]{0,1,0}{#1}}
\newcommand{\blu}[1]{\textcolor[rgb]{0,0,0}{#1}}
\newcommand{\blue}[1]{\textcolor[rgb]{0,0,0}{#1}}

\title{\LARGE{Distributed Traffic Synthesis and Classification in Edge Networks: A Federated Self-supervised Learning Approach}}

\author{Yong~Xiao, \IEEEmembership{Senior~Member,~IEEE}, Rong Xia, Yingyu Li, Guangming Shi, \IEEEmembership{Fellow, IEEE}, Diep N. Nguyen, \IEEEmembership{Senior~Member,~IEEE}, Dinh Thai Hoang, \IEEEmembership{Senior~Member,~IEEE}, Dusit Niyato, \IEEEmembership{Fellow, IEEE}, and Marwan Krunz, \IEEEmembership{Fellow, IEEE}

\thanks{*This work has been accepted at IEEE Transactions on Mobile Computing. Copyright may be transferred without notice, after which this version may no longer be accessible.

Y. Xiao was supported in part by the National Natural Science Foundation of China under grant 62071193 and the Key R \& D Program of Hubei Province of China under grants 2021EHB015 and 2020BAA002. G. Shi was  supported in part by the National Natural Science Foundation of China under grants 62293483, 61871304, and 61976169. Y. Xiao and G. Shi were supported in part by the major key project of Peng Cheng Laboratory (grant No. PCL2021A12). M. Krunz was supported by the National Science Foundation (grants 1910348, 1731164, 1813401, 2229386, and IIP-1822071) and by the Broadband Wireless Access \& Applications Center (BWAC). 

Y. Xiao is with the School of Electronic Information and Communications at the Huazhong University of Science and Technology, Wuhan 430074, China, also with the Peng Cheng Laboratory, Shenzhen, Guangdong 518055, China, and the Pazhou Laboratory (Huangpu), Guangzhou, Guangdong 510555, China (e-mail: yongxiao@hust.edu.cn).

R. Xia is with the School of Electronic Information and Communications at the Huazhong University of Science and Technology, Wuhan 430074, China (e-mail: rong\_xia@hust.edu.cn).

Y. Li is with the School of Mech. Eng. and Elect. Inform. at the China University of Geosciences, Wuhan, China, 430074 (e-mail: liyingyu29@cug.edu.cn).


G. Shi is with the Peng Cheng Laboratory, Shenzhen, Guangdong, 518055, China and is also with the School of Artificial Intelligence, the Xidian University, Xi’an, Shanxi, China, 710071, and Pazhou Lab (Huangpu), Guangdong, 510555, China (e-mail: gmshi@xidian.edu.cn).

D. Nguyen and D. T. Hoang are with the School of Electrical and Data Engineering, University of Technology Sydney, Australia (e-mail: \{diep.nguyen, hoang.dinh\}@uts.edu.au).

D. Niyato is with the School of Computer Science and Engineering, Nanyang Technological University, Singapore (e-mail: dniyato@ntu.edu.sg).

M. Krunz is with the Department of Electrical and Computer Engineering, the University of Arizona, Tucson, AZ (e-mail: krunz@arizona.edu).
}
}
\maketitle

\begin{abstract}
With the rising demand for wireless services and increased awareness of the need for data protection, existing network traffic analysis and management architectures are facing unprecedented challenges in classifying and synthesizing the increasingly diverse services and applications. This paper proposes FS-GAN, a federated self-supervised learning framework to support automatic traffic analysis and synthesis over a large number of heterogeneous datasets. FS-GAN is composed of multiple distributed Generative Adversarial Networks (GANs), with a set of generators, each being designed to generate synthesized data samples following the distribution of an individual service traffic, and each discriminator being trained to differentiate the synthesized data samples and the real data samples of a local dataset. A federated learning-based framework is adopted to coordinate local model training processes of different GANs across different datasets. FS-GAN can classify  data of unknown types of service and create synthetic samples that capture the traffic distribution of the unknown types. We prove that FS-GAN can minimize the Jensen-Shannon Divergence (JSD) between the distribution of real data across all the datasets and that of the synthesized data samples. FS-GAN also maximizes the JSD among the distributions of data samples created by different generators, resulting in each generator producing synthetic data samples that follow the same distribution as one particular service type. Extensive simulation results show that the classification accuracy of FS-GAN achieves over $20\%$ improvement in average compared to the state-of-the-art clustering-based traffic analysis algorithms. FS-GAN also has the capability to synthesize highly complex mixtures of traffic types without requiring any human-labeled data samples. 
\end{abstract}

\begin{IEEEkeywords}
Traffic classification, self-supervised learning, edge computing, federated learning, generative adversarial networks.
\end{IEEEkeywords}

\section{Introduction}
\label{Section_Introduction}

Telecommunication technologies have evolved tremendously over the past few decades, driven by innovative services and applications in a wide range of verticals \cite{6553677,6957146}. In particular, recently standardized 5G technology introduces three major use cases: eMBB (enhanced Mobile BroadBand), URLLC (Ultra Reliable Low Latency Communications), and mMTC (massive Machine Type Communications), with promises to enable novel applications including IoT \cite{7414384,7123563,XY2019MultiOperatorIoT} and autonomous vehicles \cite{6179503}. According to recent reports\cite{ITU2019Network2030,fg2019new}, the next-generation mobile technologies, e.g., B5G and 6G, are expected to further extend the application scenarios of 5G by bringing more diverse and innovative services, such as holographic-type communications, Tactile Internet\cite{XY2018TactileInternet}, semantic communications\cite{Shi2021semantic,XY2023CollaberativeJSAC}, and others\cite{XY20206GSelfLearn}.


As more services and applications are introduced and adopted at different times in different regions, network traffic analysis and management face unprecedented challenges. In particular, the diverse demands and requirements of different services significantly increase the dynamics of network traffic, making traffic prediction, network planning, resource allocation and scheduling more challenging than before \cite{5462101,XY2021AdaptiveFog}. 
Most existing solutions rely on regression or convolutional neural network (CNN)-based supervised learning to fit historical data and accordingly classify the traffic of existing services \cite{9162891}. These solutions often require a large amount of manually labeled datasets, which may not be practically feasible to obtain \cite{8620563}. Recent works  employ clustering-based unsupervised learning solutions to analyze and cluster unknown  traffic \cite{info2020traffic}. However, these solutions often suffer from limited accuracy and cannot be applied to predict or keep track of  highly mixed and evolving service traffic flows. There is a pressing need to develop intelligent traffic analysis and classification solutions that can automatically identify known services and classify/label unknown services and data samples that emerge in decentralized datasets across various locations.  

%
%

One promising solution to the above challenges is self-supervised learning, \blu{a novel algorithmic framework that can learn features and representations without requiring any labeled data samples for training machine learning (ML) models. Self-supervised learning combines the advantages of supervised-learning and unsupervised-learning by first  solving some carefully designed pretext tasks to automatically create pseudo-labels for the unlabelled data samples. It then employs supervised-learning-based downstream training tasks to construct ML  models based on pseudo-labelled dataset\cite{Liu2021SelfSupervisedLearning,wang2021self,araslanov2021self}. Due to its potential to significantly improve data efficiency and model generality, self-supervised learning is commonly believed to be one of the key enablers for the `next artificial intelligence revolution"\cite{LeCun2020SelfsupervisedLearningKeyNote}.}  

\blu{Despite their potential, most existing works on self-supervised learning focus on designing pretext tasks that utilize the attributes of images or videos, such as image orientation, gray-scale image colorization, and image jigsaw puzzle solving, none of which can be applied to analyze network traffic datasets. Also, compared to image data samples, each consists of rich (pixel-level) information with well-known representation features and attributes, network packets generally have much smaller sizes. The data streams communicated throughout the network contain highly mixed traffic types and data packets associated with different services may exhibit very similar frame patterns that are challenging to classify. There is still no commonly known attributes of data packets associated with each individual service. }

\blu{The increasingly stringent requirements on the responsiveness of smart services and the growing awareness of data protection and network security further exacerbate the above challenges. This is especially the case, considering that most existing self-supervised learning solutions are cloud-based in which every mobile device must upload its locally collected data samples to a cloud server and wait for feedback before making decisions \cite{baker2017internet,9460171}.} 
These solutions are known to suffer from long communication delay and risk of privacy leakage caused by showing raw data. To address these issues, edge intelligence has been recently introduced, which enables distributed data processing and learning over a large number of low-cost, often decentralized edge devices, e.g., mobile edge servers \cite{zhou2019edge}. 
It is considered one of the most sought-after functions in B5G/6G.

One of the challenges of implementing edge intelligence is to develop simple and effective algorithmic solutions for decentralized data learning and processing over large resource-limited wireless networks \cite{XY20206GSelfLearn}.
%
%
As an emerging distributed AI solution, federated learning (FL) allows 
collaborative construction of ML models without exposing any raw data across decentralized datasets. Integrating FL with self-supervised learning will have the potential to significantly reduce communication overhead, enhance data protection, and improve efficiency of model training. \blu{Unfortunately, conventional FL approaches often suffer from slow convergence and bias when the distributions of traffic across different datasets are highly heterogeneous and unbalanced.  In fact, the most commonly used federated learning solution, FedAvg, can only be applied when all decentralized datasets observe the same set of features. Furthermore, most existing FL solutions are based on supervised learning, which require high-quality labeled samples across all decentralized datasets \cite{lim2020federated}. Currently, there is no simple and effective self-supervised learning solution that allows joint model construction based on heterogeneous datasets.}


In this paper, we propose the federated self-supervised Generative Adversarial Networks (FS-GAN), a novel framework for distributed traffic analysis and synthesis over decentralized  datasets consisting of highly heterogeneous and unbalanced traffic types. 
\blu{FS-GAN is comprised of multiple decentralized GANs, deployed at a set of edge servers, each of which can access an exclusive set of data samples associated with a combination of local services. Different edge servers can access different combinations of services. A set of generators is deployed at the edge servers or virtual machines that offer computational functions.}  In contrast, each discriminator is implemented at an edge server and has been assigned with an exclusive right to access a local dataset requiring a certain privacy protection for local data samples. FS-GAN learns to create synthetic data samples that capture the statistical features of real data (i.e., its distribution) obtained from each individual type of services to support automatic traffic classification. \blue{Compared to a traditional GAN, FS-GAN provides three unique advantages. First, it supports collaborative model construction based on decentralized datasets by adopting a federated learning-based approach to coordinate model training at different edge servers. Second, it addresses the model collapse problem suffered by many GAN-based approaches by associating a classifier with each local datatset to classify samples generated by different generators. Third, FS-GAN does not require any labeled samples to train the model; rather, it adopts a self-supervised learning-based approach that automatically assigns different pseudo-labels to synthesized data samples generated by different generators. Compared to a traditional network data classification/clustering solution, the proposed FS-GAN trains multiple generators, each of which is capable of synthesizing data traffic for a given service. These features of FS-GAN make it possible to further improve the performance of data traffic classification, especially when traffic samples of different services are highly unbalanced, i.e., some services have much fewer data samples than others. They also make FS-GAN applicable to a wide range of scenarios and use cases that involve traffic prediction and synthesis, such as dynamic network slicing that involves multi-service traffic tracking and prediction, unknown attack detection, and prediction-based network planning.}
%
%
We conduct extensive simulations based on real-world traffic associated with ten popular services {applications} including email, FTP, video and audio chat, etc \cite{datatraffic}. Our results show that the classification accuracy of FS-GAN achieves over $20\%$ improvement, on average, compared to the state-of-the-art clustering-based traffic analysis algorithms. 

\begin{figure}[ht]
	\centering
	\includegraphics[width=9cm]{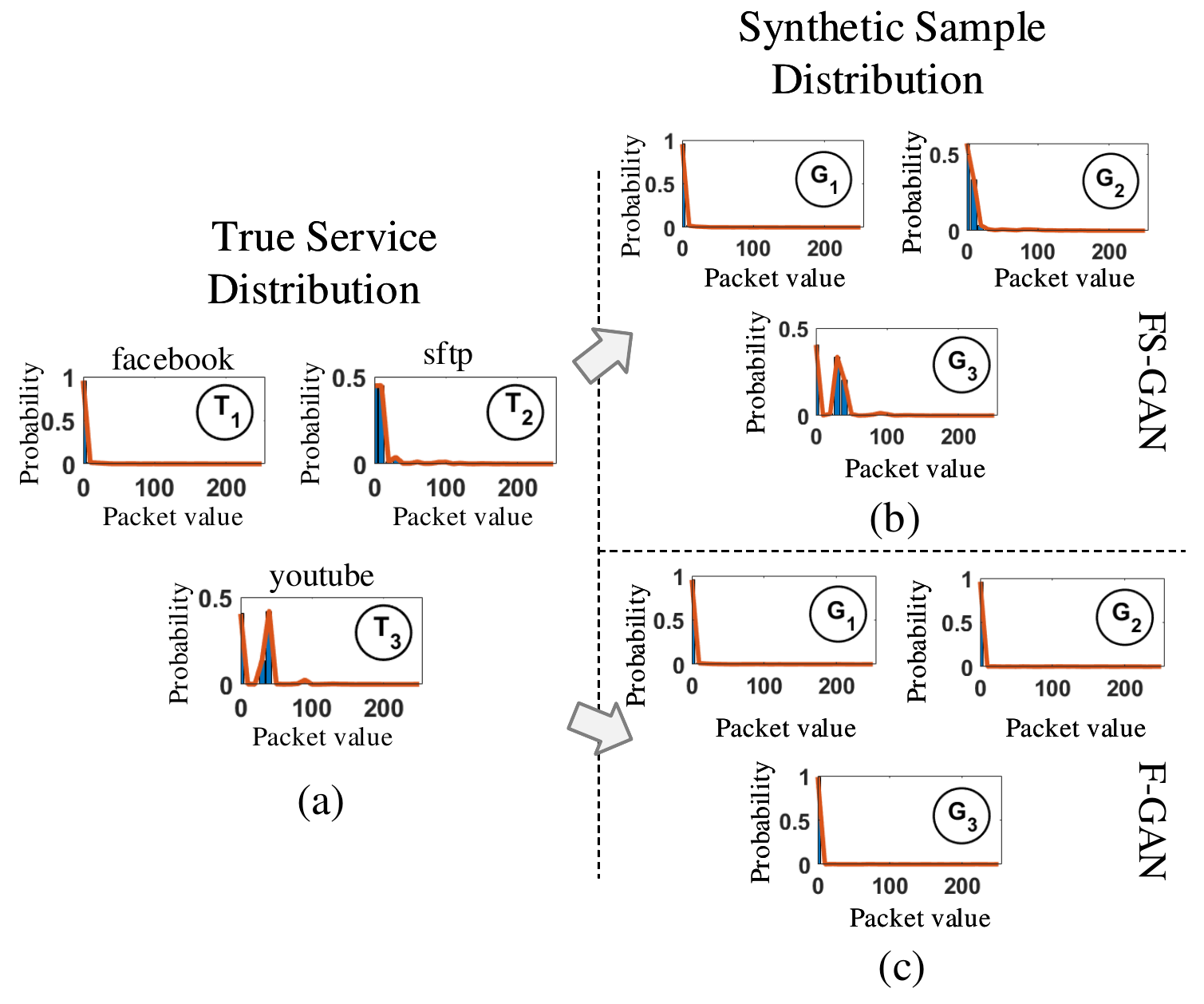}
	\vspace{-0.2cm}
	\caption{\footnotesize{Comparison between (a) the distribution of three service traffics $T_1, T_1, T_3$ measured in packet value (converted decimal value of data packet) and (b) the synthesized traffic data produced by FS-GAN and (c) the synthesized traffic produced by F-GAN.}}
\label{fig intro}
\end{figure}
\vspace{-0.2cm}

To highlight the advantages of FS-GAN, in Fig. \ref{fig intro},  we present the traffic distribution of the data packets from the real traffic data associated with 3 types of (unknown) applications (Fig. \ref{fig intro}(a)) and the distribution of synthesized traffic generated by FS-GAN after 50 rounds of training (Fig. \ref{fig intro}(b)), compared with the distribution of data produced by a straightforward extension of traditional multi-generator GAN into an FL setting, referred to as F-GAN (Fig. \ref{fig intro}(c)). \blue{It can be observed that FS-GAN is able to successfully generate synthetic data samples that match distributions of different types of applications. F-GAN, however, fails to capture data distribution differences between various services but can only generate samples with a limited diversity.} 

The main contributions of this paper are as follows:   
 \begin{itemize}
 	\item[1)]  We introduce a novel framework called FS-GAN that exploits concepts from GAN, FL, and self-supervised learning to enable automatic learning, synthesis, and classification of heterogeneous traffic over decentralized datasets.
Compared to regression-based clustering solutions, which separate data samples based on a single or a limited number of pre-selected features, FS-GAN autonomously creates synthetic samples with pseudo-labels to capture the distributions of local mixtures of traffic types and then applies supervised-learning-like solutions to further improve the classification performance. To the best of our knowledge, FS-GAN is the first network traffic synthesis and classification solution that uses self-supervised learning.

	\item[2)] We prove that the model trained by FS-GAN minimizes the difference between the distribution of the real traffic and that of the synthetic data samples created by the FS-GAN generators.
	The unique ability of FS-GAN to learn and create synthetic samples that capture the distribution of a heavily mixed but unknown traffic across decentralized datasets allows it to be applied to a range of novel applications, well beyond traditional traffic classification. We discuss some potential applications and pointed out the limitations of FS-GAN under different scenarios.
	\item[3)]  We conduct extensive simulations using real traffic datasets. Our results show that FS-GAN achieves significant improvement in both traffic classification and data synthesis.
\end{itemize}

The rest of the paper is organized as follows. In section \ref{Section_RelatedWork}, we first review recent works that are relevant to FS-GAN.  We then introduce the architecture and discuss its application scenarios in Section \ref{Section_Overview}. Details of the proposed architecture, algorithms, as well as the main theoretical results are presented in Section \ref{Section_FSGANArchitecture}. We conduct extensive simulations and evaluate the performance of FS-GAN in Section \ref{Section performance}. The paper is concluded in Section \ref{Section_Conclusion}.

\section{Related Work}
\label{Section_RelatedWork}
The solution proposed in this paper is closely related to recent progress in deep-learning-based traffic classification and FL-based data analysis and synthesis. Most related works are reviewed in this section.
\subsection{Deep-learning-based Traffic Classification}
Deep neural network (DNNs), are one of the promising solutions for network traffic classification \cite{info2020traffic,9162891,9162884,8737507,8845127,Liang2019NeuralPC,8624128,nascita2021xai,sadeghzadeh2021adversarial,aceto2021distiller}. For example, Wang \textit{et al.} proposed a hybrid neural network that combines a recurrent neural network (RNN) and a CNN to learn dynamic features of mobile  data for traffic classification\cite{9162891}. Liang \textit{et al.} introduced a deep reinforcement learning-based approach, called NeuroCuts, for packet classification\cite{Liang2019NeuralPC}.
Zhang \textit{et al.} proposed a deep learning-based traffic clustering solution that can classify packets associated with unknown classes and automatically build a new training dataset to update the classifier of unknown traffic\cite{info2020traffic}. Liu \textit{et al.} applied an RNN-based flow sequence network approach to classify encrypted traffic flows\cite{8737507}. \blu{Ensemble learning-based methods have been adopted to improve the traffic classification accuracy by Aouedi \textit{et al.}\cite{ensemble_1} and Jin \textit{et al.}\cite{ensemble_2}. }

\subsection{FL-based Network Data Analysis}
FL and its applications in network data analysis have recently attracted significant interest  \cite{yqfed,info20FLwireless,9155494,Li2019FederatedLS,Li2020FLMagazine,9162958,li2020convergence,zhu2021distributed}. FL has the potential to protect data privacy during distributed training and coordination across decentralized datasets. Thus, network Wen \textit{et al.} proposed a two-step FL framework to achieve privacy preserving model training with high communication efficiency. Wang \textit{et al.} proposed an empirically driven solution to optimize the selection of client devices that participate in model updating\cite{9155494}. Recently, FL was extended to data analysis in resource-limited networking systems. For example, Wang \textit{et al.} investigated the convergence of gradient descent-based FL solutions and optimized the tradeoff between local update and global model aggregation under a given resource constraint\cite{Wang2019AdaptiveFL}. Luo \textit{et al.} introduced a low-cost sampling-based algorithm to learn the convergence-related parameters and minimize the cost of learning time and energy\cite{Luo2020cost}. Xiao \textit{et al.} proposed a federated edge intelligence (FEI) framework for implementing FL in edge computing-assisted IoT networks with communication and computational resource constraints\cite{XY2020WCSP}. This framework has been further extended to real-time learning in edge intelligence networks\cite{XY2023TimeSensLearning}, as well as semantic communications\cite{Shi2021semantic} and semantic-aware networking systems\cite{XY2023CollaberativeJSAC,XY2022ReasononAir,XY2022ITWSSC}. \blu{Zhang \textit{et al.}\cite{analysis_2} studied the homomorphic encryption-based FL for communication and computation cost of private medical data analysis and Lu \textit{et al.}\cite{analysis_1} investigated FL-based imbalanced classification for industrial data.}     

\subsection{Self-supervised learning-based Data Analysis}
Recently, self-supervised learning was introduced as a way to improve data efficiency and model generalization \cite{araslanov2021self,li2021cutpaste,sun2020multi,9141293,wang2021self,Liu2021SelfSupervisedLearning,bengar2021reducing}.
In particular, Araslanov \textit{et al.} proposed a domain adaptation approach for semantic segmentation based on predictions produced by pseudo-supervision targets \cite{araslanov2021self}.
Li \textit{et al.} proposed a two-stage framework for anomaly detection in which a self-supervised deep representation model was learned to build a generative classifier \cite{li2021cutpaste}.
Sun \textit{et al.} introduced a novel model training algorithm for Graph Convolutional Networks (GCNs). \blu{Self-supervised model in \cite{sun2020multi} improved model generalization performance while reduce the number of required labeled samples.} \blu{Zhang \textit{et al.} proposed a self-supervised learning method with adaptive memory network to improve model generalization and enrich feature representativeness\cite{self_1}.} \blu{Shi \textit{et al.} studied a novel mask image model for self-supervised learning, where the idea of adversarial training was utilized\cite{self_2}}


Our proposed FS-GAN differs from traditional clustering-based data analysis techniques in that 
it adopts a self-supervised learning-based solution to first create synthetic samples with pseudo-labels that statistically match a mixture of real traffic types and then train an ML model using the pseudo-labeled datasets to classify and identify each individual service traffic.

\section{Architecture Overview and Application Scenarios}
\label{Section_Overview}
\subsection{Architecture Overview}
\begin{figure}[htbp]
	\centering
	\includegraphics[width=9cm]{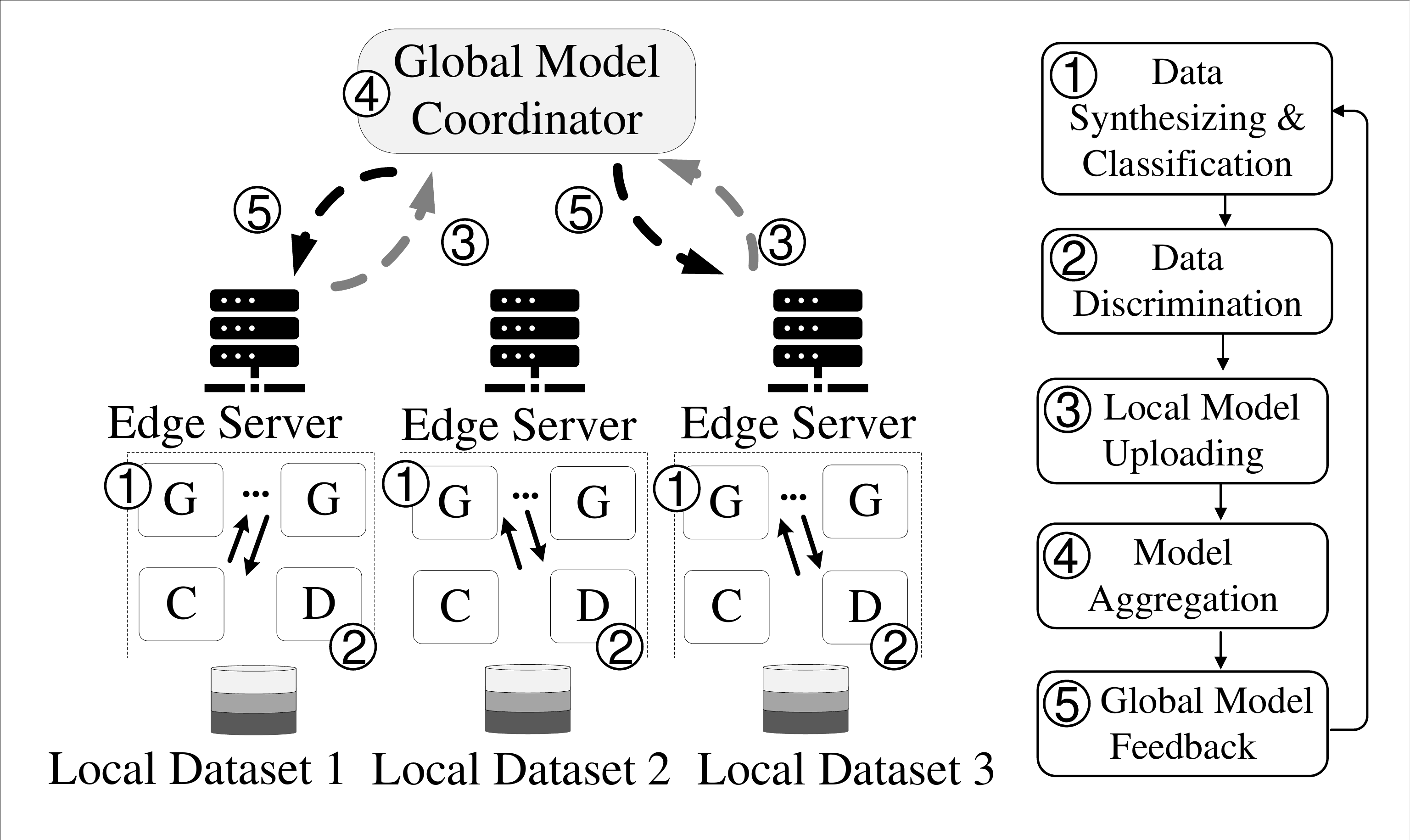}
	\centering
	\caption{\footnotesize{FS-GAN architecture, including components generators (G), discriminators (D), classifiers (C), and a global model coordinator (left figure); and the main algorithmic procedures (right figure).}}
	\label{fig FS-GAN}
\end{figure}
The main architecture, including the key components and algorithmic procedures of FS-GAN, is presented in Fig. \ref{fig FS-GAN}. FS-GAN consists of: (1) multiple generators that create synthetic samples whose statistical distribution is similar to that of real traffic in local datasets, and (2) multiple discriminators, each of which discriminator tries to differentiate the real traffic from the synthetic samples produced by the generators. 
Our proposed FS-GAN framework employs the following three procedures:

\noindent
{\bf (1)  Local Data Synthesis:}
One discriminator is deployed in each edge server with exclusive right access to a local dataset. Each discriminator is jointly trained with several associated generators. 
Although GANs, in general, have high flexibility in generating synthetic samples \cite{GAN,yu2017seqgan,chen2016infogan}, they suffer from the so-called  \textit{model collapsing problem}, where the generator learns to produce synthetic samples with extremely limited diversity, e.g., samples that only capture the data distribution of a single traffic type or a fixed mixture of a given set of traffic types.
To tackle this problem, we borrow the idea of multi-generator GANs \cite{MixGAN,MGAN,zhang2018Stackelberg} and assign a pseudo-label to each generated sample to indicate which generator the sample came from. As will be proven later, a generator that is locally trained with a discriminator is able to create synthetic data samples that match the data distribution of the real traffic in the local dataset.


\noindent
{\bf (2) Global Model Coordination:}
The local models trained by the discriminators and generators will be periodically uploaded to a global model coordinator. The coordinator can be deployed at one of the edge servers or at the cloud data center. One straightforward implementation of model coordination is to adopt FL solutions\cite{Li2019FederatedLS,Li2020FLMagazine}, in which the local model parameters (of generators and/or discriminators) are aggregated, e.g., averaged in FedAvg algorithm\cite{FedAvg}, to form an updated global model that is then broadcasted to edge servers. Note that the number of model parameters  from local edge servers will affect the communication efficiency. In this paper, we consider two model coordination schemes:
\begin{itemize}

\item[{\bf (i)}] {\bf Coordination Scheme I (C-I):} Both locally trained discriminators and the associated generators are uploaded to the global coordinator.

\item[{\bf (ii)}] {\bf Coordination Scheme II (C-II):} Only discriminators are uploaded and aggregated by the coordinator.
\end{itemize}

C-II requires less model-related data to be exchanged between edge servers and the coordinator, resulting in a higher communication efficiency. However, as will be shown later in this paper, C-I has faster convergence than C-II, 
which in some cases compensates for the extra communication overhead.

In the local data synthesis procedure, each discriminator only determines whether the data samples belong to a real traffic or are synthetic data produced by generators. After the global model coordination, all the discriminators should be able to differentiate real data in \textit{any} local dataset from synthetic data generated by \textit{any} generator. Similarly, the generators will be able to capture richer features across different local datasets. To address the model collapsing problem, we introduce a classifier in each local datatset to classify samples from different generators. As shown in Section \ref{Section_FSGANArchitecture}, this classifier increases the divergence of the synthetic data distributions of different generators, hence alleviating the model collapsing problem of traditional GANs.

\noindent
{\bf (3) Self-Supervised Learning:}
%
The aforementioned trained and updated classifier creates pseudo-labels for synthetic samples generated by different generators. As the generators become more capable of producing close-to-real synthetic data, the classifier associated with each discriminator will naturally support data classification and self-labeling of real traffic that arrive in the future. We show that the classifier actually shares the same model parameters as the locally trained discriminator, and therefore does not require any extra effort for model training.

We present a theoretical analysis that proves the effectiveness of the FS-GAN framework. In particular, we prove that FS-GAN possesses two important properties: (1) the difference between the distributions of the mixture of the synthetic data generated by generators and that of the real traffic data distribution is minimized, and (2) the \mbox{JSD} divergence of the distributions of synthetic data samples generated by different generators is maximized. Based on these properties, we prove the main result of this paper, namely, the synthetic data samples generated by each individual generator captures the real traffic distribution of each individual traffic type, and that the classifier can differentiate all the synthetic samples coming from different generators, if the distributions of the traffic data associated with different services are separable. 

\blue{Compared to existing self-supervised learning solutions, FS-GAN provides the following unique advantages that make it more suitable for network traffic classification problems. First, FS-GAN can be applied to a large network with highly decentralized and imbalanced traffic distribution with data privacy protection requirements. Second, FS-GAN does not require any pre-labeling of data, which makes it suitable for many of today’s networking systems that involve a large volume of unlabeled network traffic datasets that are prohibitively expensive to label. Third, the data synthesizing capability of FS-GAN makes it possible to be applied in many emerging networking services that involve traffic prediction and synthesis.}


\subsection{Examples of Application Scenarios}

\blue{Compared to traditional traffic classification solutions, FS-GAN adopts a fundamentally different approach by first training multiple generator networks to generate synthetic samples that capture the distribution of real traffics associated with an individual service and then applying the parameters of the already trained models for classifying traffics of different services. This uniqueness of FS-GAN makes it applicable to a range of new application scenarios and use cases, including the following}:

{\bf 1) Dynamic Network Slicing in 5G/B5G:}
5G NR release 16 \cite{3GPP} introduced the concept of network slicing, which focuses on 
isolating  and preserving partitions of network resources and functions, commonly referred to as \textit{slices}. These slices are tailored  and orchestrated to support applications with diverse QoS requirements. One of the key challenges in network slicing is now to keep track of the traffic demands of potentially unknown applications, so the appropriate amount of resources can be reserved while meeting the needs of these applications. FS-GAN can be directly applied to classify and keep track of the needs of different services from a highly mixed traffic. It can also assist the network slicing manager in identifying newly observed service traffic types that do not have a sufficient amount of pre-labeled data.

{\bf 2) Unknown Attack Detection: }
An important application scenario for traffic classification is attack detection. Existing solutions mostly focus on detecting known attacks. Detecting unknown attacks is a notoriously difficult problem\cite{attack_detection}. FS-GAN has the potential to identify unknown attacks and simulate various attack scenarios as well as their mixtures.

{\bf 3) Traffic Prediction-based Network Planning: }
Because the final model obtained by FS-GAN is capable of generating synthetic data samples that are statistically similar to the real data of various types of emerging traffic, FS-GAN can be applied to predict future traffic trends and assist in planning the network infrastructure to better cope with anticipated needs.

\section{FS-GAN Design and Theoretical Analysis}
\label{Section_FSGANArchitecture}

\begin{figure*}[htbp]
	{
		\centering
		\includegraphics[width=14cm]{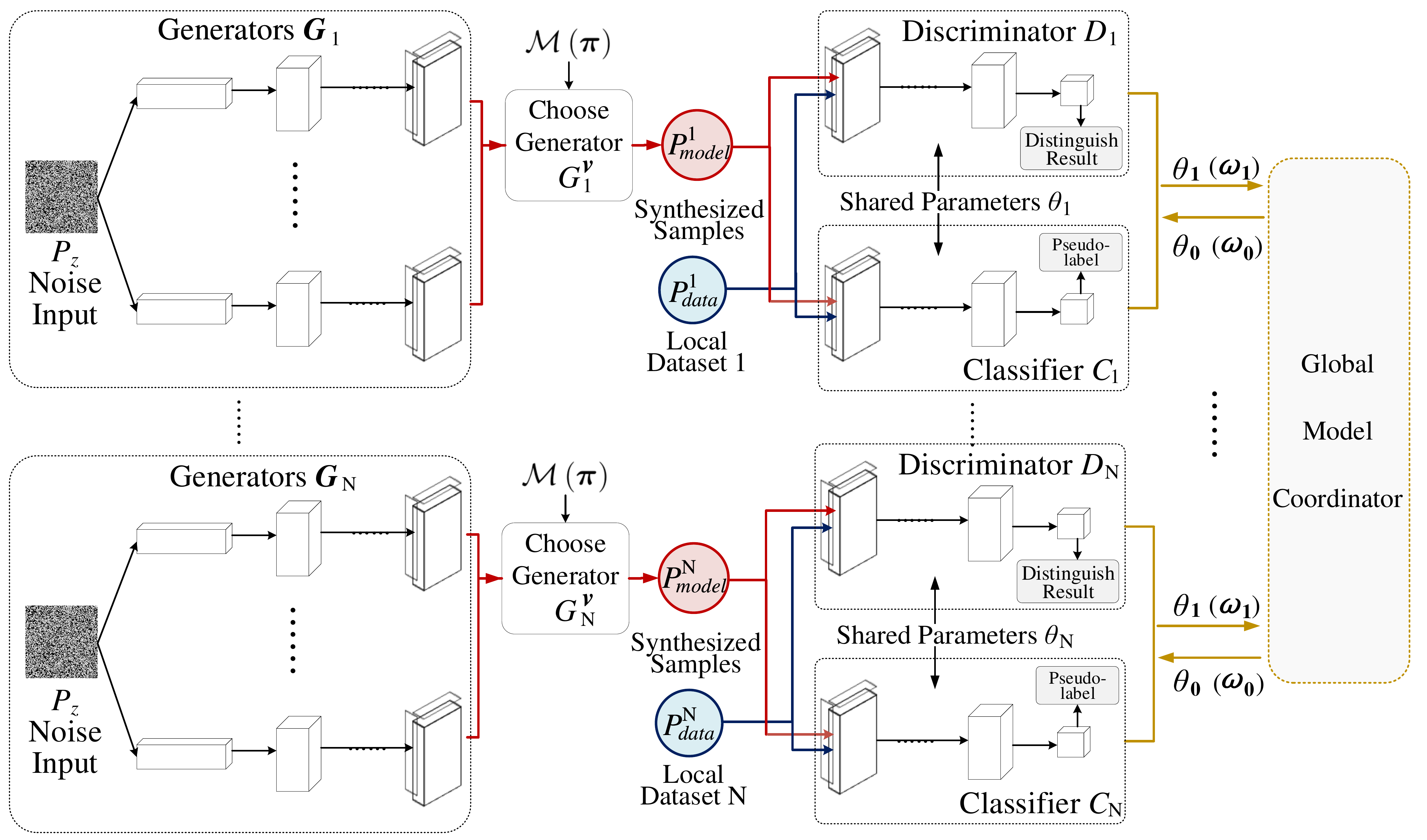}
	}
	\centering
	\caption{\footnotesize{Detailed architectural components of FS-GAN}}
	\label{fig FMGAN}
\end{figure*}

In this section, we present the algorithmic details
of the main procedures of FS-GAN (both C-I and C-II): local data synthesis, global model coordination, and self-supervised learning. The detailed architecture is shown in Fig. \ref{fig FMGAN}. Theoretical analysis is presented at the end of this section.

\begin{table}[htb]
    \centering
\blu{
\caption{List of Notation}
    \begin{tabular}{c|c}
    \hline
    Notation & Description  \\
    \hline
    $d$ & the index of local datasets \\
    \hline
    $D$ & discriminator of FS-GAN \\
    \hline
    $G$ & generator of the FS-GAN \\
    \hline
    $P$ & distribution of data \\
    \hline
    $z$ & noise of Gaussian distribution \\
    \hline
    $m$ & index of generator \\
    \hline
    $\mathcal{L}$ & loss function \\
    \hline
    $x$ & real data sample \\
    \hline
    $\hat{x}$ & synthetic data sample \\
    \hline
    $\lambda$ & diversity hyper-parameter \\
    \hline
    $\omega$ & parameters of generator \\
    \hline
    $\theta$ & parameters of discriminator and classifier \\
    \hline
    \end{tabular}
}
    \label{notations}
\end{table}

\subsection{Local Data Synthesis}
First, we consider local-model training of a single generator at the $d$th local dataset, $d=1, ..., N$.
As mentioned before, the main objective of this procedure is to train a model to generate synthetic data samples that captures the distribution of real traffic of a local dataset.
Let $G_d$ and discriminator $D_d$ be, respectively, the generator and discriminator associated with the $d$th dataset.
Training GANs can be formulated as a minimax game between $G_d$ and $D_d$ \cite{GAN}. The generator and the discriminator are often DNNs. Let $P_{data}$ be the distribution of the real dataset and suppose that $\bx$ is drawn from $P_{data}$, i.e., we write as $\bx\sim P_{data}$. The generator $G_d$ aims to train a multi-layer neural network to map its input variables. Initially, a random noise $\bz$ is generated to synthesize data samples whose distribution is denoted as $P_{\bz}$. 
Without loss of generality, we abuse the notation and use $G_d(\bz)$ to denote the trained generator's output given training input $\bz$. Meanwhile, the discriminator $D_d$ tries to distinguish real data samples from synthetic samples produced by the generator. We use $D_d(\cdot)$ to denote the output of discriminator, which represents the probability that the discriminator decides the given input sample comes from the real dataset rather than being synthesized by the generator. More formally, we can express the adversarial training process between a single generator and a single discriminator via the following minimax problem:
\begin{equation}\label{eq GAN}
\begin{aligned}
\min_{G_d}\max_{D_d} \mathbb{E}_{\bx\sim P_{data}}[\log D_d(\bx)] +\mathbb{E}_{\bz\sim P_{z}}[\log\left(1-D_d(G_d(\bz))\right)].
\end{aligned}
\end{equation}


The key idea of FS-GAN is to use multiple generators ${\boldsymbol G}_d=\{G_d^1, G_d^2, ..., G_d^M\}$, rather than a single one to jointly train a model that can capture the distribution of a mixture of different traffic types in a given local dataset $d$. 
To address the issue of model collapse, we introduce a classifier network $C_d$ that classifies synthetic data produced by the generators. More specifically, the classifier first generates a pseudo-label for each synthetic data sample, indicating which generator it came from. It then minimizes its loss to encourage generators to create samples that are different from  each other. As we prove later in this section, by employing $C_d$, each of the generators will synthesize data samples that represent a specific type of traffic, provided that the differences between the distributions of different traffic types are sufficiently large.

Now we focus on the training process for the multi-generator local model. We use the subscript to denote the index of the local dataset and the superscript for the generator index.
Suppose a set ${\boldsymbol G}_d=\{G_d^1, G_d^2, ..., G_d^M\}$ of $M$ generators has already been assigned to the discriminator $D_d$, where $d\in\{1,2,...,N\}$ represents the discriminator index. Each discriminator $D_d$ also interacts with a local dataset $X_d$, and each generator $G^{m}_d, m\in \{1,2,...,M\}$, maps the input noise $\bz$ to $\hat{\bx} = G^{m}_d(\bz)$. The classifier $C_d$, is co-trained with ${\boldsymbol G}_d$ and $D_d$. Let $P_{G^{m}_d}$ be the distribution of the synthetic samples produced by $G^{m}_d$. We now describe the rule for each generator to interact with the discriminator. We set  
an index vector $\boldsymbol v$, randomly drawn from a predefined multinomial distribution $\boldsymbol v \sim {\cal M} \left( \boldsymbol{\pi} \right)$ where $\boldsymbol{\pi} = \langle  {\pi}_{\boldsymbol v} \rangle_{\boldsymbol v\in {\boldsymbol G}_d}$ is the weighting coefficient of the distribution mixture. By setting $G^{\boldsymbol v}_d\triangleq\{G^i_d: i\in\ \boldsymbol v\}$ as the  subset of  ${\boldsymbol G}_d$, we can convert the multi-generator GANs into a standard GAN with a generator vector ${G}^{\boldsymbol v}_d$ and discriminator $D_{d}$.
%
We use $P_{data}^d$ to denote the real distribution of dataset $X_d$. We can then formulate the model of local data synthesis with a multi-generator as the following minimax game:
\begin{equation}\label{eq localMixGAN}
\begin{aligned}
\min_{{\boldsymbol G}_d,C_d}\max_{D_d} \quad & \; \mathbb{E}_{\bx\sim P_{data}^d}[\log D_d(\bx)] \\
&+ \sum_{m=1}^M\mathbb{E}_{\bz \sim P_z}[\log\left(1-D_d(G^{m}_d(\bz))\right)]\\
& -\lambda\sum_{m=1}^M \left( \pi_{m} \mathbb{E}_{\hat{\bx}\sim P_{G^{m}_d}}[\log C^{m}_d(\hat{\bx})] \right)
\end{aligned}
\end{equation}
where $C^{m}_d(\hat{\bx})$ is the output of classifier $C_d$, specifying the probability that data sample $\hat{\bx}$ is generated by $G^{m}_d$. 
The last term in (\ref{eq localMixGAN}) is a standard softmax loss in a multi-classification setting \cite{softmax}. In other words, minimizing $C_d$ according to (\ref{eq localMixGAN}) under fixed $D_d$ minimizes the entropy for $C_d$, which leads to increased divergence among generators. We introduce a diversity hyper-parameter $\lambda > 0$ to specify the degree that the classifier can influence the generators. We compare the model performance under different $\lambda$ in Section \ref{Section performance}.

Let $\omega_d$ be the model parameters of the generators ${\boldsymbol G}_d$ associated with discriminator $D_{d}$, and let $\omega_d^{m}$ be the model parameters of the generator ${G^{m}_d}, m\in\{1,...,M\}$.
Classifier $C_d$ and discriminator $D_d$  share the same model parameters except for the last layer. Let $\theta_d$ be the  model parameters shared between the classifier and the discriminator $D_{d}$. Note that the last layer of $C_d$ outputs an automatically learned pseudo-label for each input data sample, while the last layer of $D_d$ outputs a binary result. Similar to a standard GAN, we alternatively learn $\omega_d$ and $\theta_d$ using the stochastic gradient descend (SGD) method. The local data synthesis framework is illustrated in Fig. \ref{fig FMGAN}.

Let $B$ be the selected mini-batch size during the training process. At the beginning of each local training iteration, each discriminator $D_d$ will sample a mini-batch of $B$ data points $(\bx_d^{(1)}, \bx_d^{(2)}, ..., \bx_d^{(B)})$ from the real data distribution $P_{data}^d$.
A mini-batch of $B$ synthetic data samples $(\hat{\bx}_d^{(1)}, \hat{\bx}_d^{(2)}, ..., \hat{\bx}_d^{(B)})$ will also be created by a mixture of generators {$G_d^v$} in ${\boldsymbol G}_d$.
This mixture can be generated according to a predefined sampling probability $\pi_m$.
During each training iteration, we update $C_d$, $D_d$, and ${{G}^{\boldsymbol v}_d}$ along the gradients of their loss functions. According to (\ref{eq localMixGAN}), we can write the loss functions under the aforementioned settings as follows:
\begin{subequations}\label{eq_SGD}
	\begin{align}
	&\mathcal{L}({C_d},\hat \bx_d)  =-\frac{1}{B}\sum_{b=1}^{B}\log C^{m}_d\left(\hat{\bx}_d^{(b)}\right)  \label{eq_SGDC}                                  \\
	&\mathcal{L}({D_d},\bx_d, \hat \bx_d)   =-\frac{1}{B}\sum_{b=1}^{B}\left(\log D_d\left(\bx_d^{(b)}\right) \right.\nonumber \\
	&\;\;\;\;\;\;\;\;\;\;\;\;\;\;\;\;\;\;\;\;\;\;\;\;\;\; \left.+ \log\left(1-D_d\left(\hat{\bx}_d^{(b)}\right)\right)\right)  \label{eq_SGDD} \\
	&\mathcal{L}({{G}^{\boldsymbol v}_d},\hat \bx_d)  =-\frac{1}{B}\sum_{b=1}^{B}\left(\log D_d\left(\hat{\bx}_d^{(b)}\right)+\lambda\log C^{m}_d\left(\hat{\bx}_d^{(b)}\right)\right)\label{eq_SGDG}
	\end{align}
\end{subequations}
We summarize the detailed procedure in Algorithm \ref{Algorithm 1}.

\begin{algorithm}[h]
	\caption{Local Data Synthesis}\label{Algorithm 1}
	{\bf Input:}  Local dataset $X_d$; noise distribution $P_{\bz}$; sampling probability $\pi_{m}$;\\
	{\bf Output:} Updated local parameter $\omega_d$ generators ${\boldsymbol G}_d$; updated local parameter $\theta_d$ for discriminator $D_d$ and classifier $C_d$;\\
	{\bf Initialize:} $\lambda > 0$; mini-batch size $B$; network parameters $\omega_d$ and $\theta_d$; maximum number of Iterations $I$;
	
	\quad {\bf  for} each iteration $i \leq I$
	\begin{itemize}
		\item[] Generator ${\boldsymbol G}_d$ {\bf do}:
		\item[] \begin{enumerate}
			\item Select generators ${G}^{\boldsymbol v}_d$ according to $\pi_m$;
			\item Each generator  $ G_m^d\in {G}^{\boldsymbol v}_d $ simultaneously  {\bf do}:
			\begin{enumerate}
				\item[$\bullet$] Sample a noise input variable $\bz$ from $P_{\bz}$;
				\item[$\bullet$] Synthesize a batch of data sample $\hat{\bx}_d^{(b)}, b\in\{1, 2, ..., B\}$;
			\end{enumerate}
			\item Send the synthesize data sample $\hat{\bx}_d^{(b)}$ to the discriminator $D_d$ and  classifier $C_d$;
			\item Update  $\omega_d$ by ascending along its gradient of loss function: $\nabla_{\omega_d}\mathcal{L}({{G}^{\boldsymbol v}_d})$;
		\end{enumerate}
		\item[] Discriminator $D_d$ {\bf do}:
		\item[] \begin{enumerate}
			\item Sample a batch of data $\bx_d^{(b)}, b\in\{1, 2, ..., B\}$ from local dataset $X_d$;
			\item Distinguish $\bx_d^{(b)}$ and $\hat{\bx}_d^{(b)}$;
			\item Update $\theta_d$ by descending along their gradients of loss functions: $\nabla_{\theta_d}\left(\mathcal{L}({C_d})+\mathcal{L}({D_d}) \right)$;		
		\end{enumerate}
		\item[] Classifier $C_d$ {\bf do}:
		\item[] \begin{enumerate}	
			\item Classify the synthesized samples $\hat{\bx}_d^{(b)}$ with respect to  different generators;
			\item Update $\theta_d$ by descending along their gradients of loss functions:  $\nabla_{\theta_d}\left(\mathcal{L}({C_d})+\mathcal{L}({D_d}) \right)$;		
		\end{enumerate}
		\item[] $i = i+1$;
	\end{itemize}
	\quad	{\bf Return} locally trained parameters
\end{algorithm}



\subsection{Global Model Coordination}
In the previous section, the generators associated with a local discriminator produce  synthetic data samples that capture the distributions of different traffic types in the local dataset. However, different local datasets may have different mixtures of traffic. It is therefore desirable to train a global model that can capture the diverse traffic across all local datasets. We adopt the FL-based framework to coordinate the local model training procedures without requiring any exchange of local datasets. One of the key ideas in FL is to coordinate different local model training procures via their model parameters. In one of the popular FL solutions, called FedAvg, each edge server periodically uploads its locally trained models to a global coordinator. The uploaded models are then averaged to form an updated global model, which will be broadcasted to all edge servers.


Formally, let $\Omega'$ be the subset of discriminators in coordination scheme C-II (or discriminator-and-generator pairs in scheme C-I) that has been selected for uploading the local models. Note that $\Omega' \subseteq \Omega$.
Let $\omega_0$ and $\theta_0$ be the corresponding global model parameters, and let $p_d$ be the weight of the $d$th local dataset such that $p_d\geq 0$ and $\sum_{d=1}^{N}p_d = 1$.
We can write the global model coordination procedure as follows:
\begin{eqnarray}\label{eq_FL_Formulation}
\left\{ {\begin{array}{*{20}{l}}
\omega_0 =  {\sum\limits_{d \in \Omega '} {{p_d}} \omega _d \;\mbox{ and }\; \theta_0 = {\sum\limits_{d \in \Omega '} {{p_d}} \theta _d,}}&{\mbox{if Scheme {C-I}}},\\
\theta_0 = {\sum\limits_{d \in \Omega '} {{p_d}} \theta _d,}&{\mbox{if Scheme {C-II}}}.
\end{array}} \right.
\end{eqnarray}

\begin{algorithm}[h]
	\caption{Global Model Coordination (Scheme C-I)}\label{Algorithm 2_FSGAN1}
	{\bf Input:} Locally trained parameter $\theta_d$ of  $D_d$ and  $C_d$; locally trained parameter $\omega_d$ of  $\boldsymbol G_d$;
	
{\bf Output:}   Aggregated parameter $\theta_0$ for the global discriminator $D_0$ and classifier $C_0$; aggregated parameter $\omega_0$ for the global generators $\boldsymbol G_0$
	
{\bf Initialize:} The number of local datasets $N$; maximum number of communication rounds $J$;

\quad {\bf  for} the number of communication rounds $j \leq J$
\begin{itemize}
\item[] The global model coordinator  {\bf do}:
\item[] \begin{enumerate}			
          \item  Calculate the aggregated parameter $\theta_0$ and $\omega_0$ according to equation (\ref{eq_FL_Formulation}):\\
          $\theta_0$,~$\omega_0~\leftarrow~$\textit{Local Data Synthesis}($X_d$, $\theta_d$, $\omega_d$)
          \item Broadcast  $\theta_0$ to and $\omega_0$ to local discriminator, classifier and generator, respectively;
        \end{enumerate}
\item[] Each local model  {\bf do}:
\item[] \begin{itemize}
          \item[]  Initialize local parameter $\theta_d$ and $\omega_0$ by:
          \item[] \quad $\theta_d~ \leftarrow~\theta_0$;~$\omega_d~ \leftarrow~\omega_0$;
        \end{itemize}
\item[] $j = j+1$;
\end{itemize}
\quad	{\bf end for}	
\end{algorithm}

\begin{algorithm}[h]
	\caption{Global Model Coordination (Scheme C-II)}\label{Algorithm 2_FSGAN2}
	{\bf Input:}  Locally trained parameter $\theta_d$ of  $D_d$ and  $C_d$;\\
	{\bf Output:}   Aggregated parameter $\theta_0$ for the global discriminator $D_0$ and classifier $C_0$;\\
	{\bf Initialize:} The number of local datasets $N$; maximum number of communication rounds $J$;
		
	\quad {\bf  for} the number of communication rounds $j \leq J$
	\begin{itemize}
		\item[] The global model coordinator  {\bf do}:
		\item[] \begin{enumerate}			
			\item  Calculate the aggregated parameter $\theta_0$ according to equation (\ref{eq_FL_Formulation}b):\\
			$\theta_0~\leftarrow~$\textit{Local Data Synthesis}($X_d$, $\theta_d$)
			\item Broadcast  $\theta_0$ to local discriminator and classifier;
		\end{enumerate}
		\item[] Each local model  {\bf do}:
		\item[] \begin{itemize}
			\item[]  Initialize local parameter $\theta_d$ by:
			\item[] \quad \quad \quad $\theta_d~ \leftarrow~\theta_0$;
		\end{itemize}
		\item[] $j = j+1$;
	\end{itemize}
	\quad	{\bf end for}
\end{algorithm}

\subsection{ Self-Supervised Learning}
As mentioned in the previous procedures, the classifier shares the same network parameters with the local discriminator. Therefore, this classifier can be updated in the global model coordination procedure 
to classify and create pseudo-labels for the synthetic samples according to their corresponding generators. This classifier will also be able to classify the real data associated with all traffic types in all datasets.
Since this classifier is a part of the training process of FS-GAN and is trained based on the pseudo-labeled synthetic datasets, it does not introduce any extra training efforts. 

\blu{
\subsection{Implementation Complexity}
Since FS-GAN reuses the model parameters of discriminators and therefore does not have to introduce any new computational complexity, compared to the traditional federated learning implementation of GAN. In other words, the complexity of implementing FS-GAN is closely related to two types of implementation cost: communication and computational cost. The computational complexity of FS-GAN mainly includes the training of a set of local generative and discriminative models in parallel based on SGDs, i.e., the total number of local SGD rounds performed by edge servers are given by $O(MKT)$. The communication cost depends on the model size and the number of coordination rounds. In other words, FS-GAN introduces novel capabilities and address the issues of traditional federated GAN without introducing any extra computational and communication complexity.}  

\subsection{Theoretical Analysis}
We analyze FS-GAN from the following two aspects: data synthesis ability and  classification accuracy across local models.
Consistent with our previous notation, we use $P_{data}^d$ to denote the mixture distribution of real data samples in dataset $X_d$, and use $P_{model}^d$ to denote the distribution of the combination of synthetic data samples generated by generators $G^1_d, G^2_d, ..., G^M_d$. First, we provide the optimal solution for classifier $C_d$ in Equation (\ref{eq localMixGAN}).
\begin{proposition} 
For a set of generators ${\boldsymbol G}$ and for a given sampling probability $\pi_m$, the optimal distribution learned by classifier ${C_{d}}^*(\bx;\theta_d)$ has the following form:
	\begin{equation}\label{eq prop1}
		\begin{aligned}
		{C^m_{d}}^*(\bx;\theta_d)&=\dfrac{\pi_m P_{G^m_d}(\bx)}{\sum_{i=1}^{M}\pi _iP_{G^i_d}(\bx)}, m \in\{1, 2, ..., M\}
		\end{aligned}
	\end{equation}
\end{proposition} 
\begin{IEEEproof}
See Appendix \ref{apend1}.
\end{IEEEproof}

The result of
the output of the optimal classifier ${C_{d}}^*(\bx;\theta_d)$ can be seen as a weighted normalization of different synthesized sample distributions.

We follow a commonly adopted setting and use Jensen Shannon Divergence (JSD) \cite{JSD} to quantify the difference between two distributions.
For the optimal generator ${{\boldsymbol G}_d}^*(\bx;\omega_d)$, we have the following proposition:
\begin{proposition}
Given the optimal discriminator and classifier, the objective of generators is to minimize:\\
	\begin{eqnarray}\label{eq prop2}
	{{\boldsymbol G}_d}^*(\bx;\omega_d)&=&\mathop{\arg\min}\limits_{G}(2\cdot \mbox{JSD}(P_{data}^d\|P_{model}^d) \\
	&&-\lambda \cdot \mbox{JSD}_{\pi_{1},\pi_{2},...,\pi_{M}}(P_{ G^{1}_d},P_{ G^{1}_d},...,P_{ G^{M}_d}). \nonumber
	\end{eqnarray}
\end{proposition}
\begin{IEEEproof}
See Appendix \ref{apend2}.
\end{IEEEproof}


It can be observed that the two terms on the right-hand-side of (\ref{eq prop2}) correspond, respectively, to the difference between distributions $P_{data}^d$ and $P_{model}^d$, and the inverse of the difference among the synthetic data generated by different generators. In other words, minimizing these two terms directly results in minimizing the difference between $P_{data}^d$ and $P_{model}^d$ while maximizing the differences among the generators.

Before presenting the main theorem, we make the following assumption:
%

\begin{assumption}
	The real data distribution $P_{data}^d$ of dataset $X_d$ can be written in the form:
	\begin{eqnarray}\label{eq assumption1}
		P_{data}^d(\bx)&=&\sum_{m=1}^{M}\pi_m p_d^m(\bx),
	\end{eqnarray}
	where for any given $\bx$, if $p_d^m(\bx)>0$ for $m\in \{1,2,...,M\}$, then $m' \neq m, ~ p_{d}^{m'}(\bx)=0$.
\end{assumption}

Equation (\ref{eq assumption1}) means that the data distribution of data samples is a mixture of $M$ separable distributions given by $p_d^m(x)$ for $m=1, 2, ..., M$. 
We can prove the following theorem:

\begin{theorem}
\label{Theorem_main}
Suppose that Assumption 1 holds. At the equilibrium point of the minimax game defined in equation (\ref{eq localMixGAN}), the optimal classifier and generators satisfy the following equations:
	\begin{subequations}\label{eq theorem1}
		\begin{align}
& {C_d^{m}}^*(\bx) \rightarrow
		\begin{cases}
		 1, \text{if}~  x\sim P_{G_d^m},\\
	     0, \text{if}~  x\sim P_{G_d^{m'}},~{\rm for }~m'\neq m, 
		\end{cases}	
		\label{eq theorem1(a)}	\\
& P_{G_d^m}^*(\bx)=p_d^m(\bx), ~\forall m=1, 2, ..., M,\label{eq theorem1(b)}\\
& P_{model}^d(\bx)= P_{data}^d(\bx). \label{eq theorem1(c)} 
		\end{align}
	\end{subequations}
\end{theorem}
\begin{IEEEproof}
See  Appendix \ref{apend3}.
\end{IEEEproof}

We can observe from (\ref{eq theorem1(a)}) that the classifier can correctly differentiate synthetic samples produced by different generators, i.e., the probability for the classifier to identify the correct generator $G_d^{m}$ that produces each given synthetic samples $x$ approaches one.
As shown in (\ref{eq theorem1(b)}) and (\ref{eq theorem1(c)}), 
synthetic samples produced by optimal generator ${G_d^m}^*$ follow the same distribution as $p_d^m$. Also, all the generators weighted by $\pi_m$ together synthesize samples that match the same data distribution of that of the real dataset.

In cases where Assumption 1 is not satisfied, Theorem 1 can be considered as an upper-bound for the local data synthesis and classification performance.
Existing works as well as our own experiments show that in many practical scenarios, even if Assumption 1 does not hold, FS-GAN can still create high-quality synthesized data with high classification accuracy performance.
We will come back to this issue in Section \ref{Section performance}.

So far, we have analyzed the optimal solution of the local model. Next, we focus on the convergence of the global model.
In the global model coordination procedure, we use Federated Averaging (\textit{FedAvg}) \cite{FedAvg}, a widely popular algorithm in federated learning, to periodically aggregate local models. 
More specifically, suppose that each participating local model performs $I$ local updates after receiving the latest global model. Let $T$ be the total number of local iterations performed by each client. Let $B$ be the mini-batch size. Let $\mathcal{L}_d(\boldsymbol{\varepsilon},\xi_t^d), d\in\{1, 2, ..., N\}$, be the loss function of local model in $t$th iteration with network parameters $\boldsymbol{\varepsilon}$ and a mini-batch of $B$ samples $\xi_t^d$. We define $\mathcal{L}_d(\boldsymbol{\varepsilon})$ as the mean value of $\mathcal{L}_d(\boldsymbol{\varepsilon},\xi_t^d), t=1,...,T$, i.e., we have $\mathcal{L}_d(\boldsymbol{\varepsilon}) = \mathbb{E}[\mathcal{L}_d( \boldsymbol{\varepsilon},\xi_t^d)]$.
Consider the following optimization model:
\begin{equation}
\begin{aligned}
	\min_{\boldsymbol{\varepsilon}} \{\mathcal{L}( \boldsymbol{\varepsilon}) \triangleq \sum_{d=1}^{N} p_d\mathcal{L}_d( \boldsymbol{\varepsilon}) \}.\label{distributed_opt_model}
\end{aligned}
\end{equation}

Theoretical guarantees of (\ref{distributed_opt_model}) can be obtained by following the same approach as \cite{fedgan,Zhao2018FederatedLW,li2020convergence}. 

\blu{
We can observe that Theorem \ref{Theorem_main} has proved that the proposed FS-GAN has addressed the two fundamental issues of federated learning and self-supervised GAN. In particular, compared to the federated learning which requires all the decentralized datasets to share the same combinations of data features, FS-GAN allows edge servers with different combinations of service traffics to jointly construct a shared model that can identify and synthesize service traffics of all the edge servers. Also, FS-GAN addresses the model collapse problem of GAN by reuse the model trained by the discriminator. Furthermore, as described in Section 4.4, FS-GAN does not introduce any new components or increase the complexity, compared to traditional federated GAN. This further justifies the flexibility and practicality of FS-GAN in practical networking systems.
}

\section{Performance Evaluation}
\label{Section performance}

\blu{In this section, we conduct extensive experiments using real-world traffic datasets to verify two key capabilities of FS-GAN proved in Theorem 1 including classification of unknown service traffics and synthesis of different types of service traffics. In the rest of this section, we first introduce the setup of our simulations in Sections 5.1 and then present detailed simulation results to evaluate the classification performance and traffic data synthesis performance of FS-GAN in Sections 5.2 and 5.3, respectively. }

\subsection{Simulation Setup}

To evaluate the performance of FS-GAN, we consider real-world dataset, 
“VPN-nonVPN dataset” (ISCXVPN2016)\cite{datatraffic}, to {evaluate} a highly mixed traffic data flow consisting of multiple types of unknown services. Within this dataset, we select data samples associated with 10 popular  services, summarized in Table \ref{tb options}. As our focus is on classification of valid information, we remove the PHY and MAC headers from the packets and limite the length of each payload to the same size of 2500 bytes so as to achieve a proper trade-off between the training efficiency and classification accuracy.
\begin{table}[htbp]\scriptsize
	\centering
	\caption{Dataset Used for Evaluation of FS-GAN}
	\begin{tabular}{|c|c|c|c|}
		\hline
		Service& $\#$ of Packets& Service& $\#$ of Packets\\
		\hline
		Email& 32,566& Youtube &12,738\\
		\hline
		ftps (upload)& 47,795& sftps (upload)&107,234\\
		\hline
		SCP (download)& 85,018& Skype Video& 140,569\\
		\hline
		Facebook Audio& 91,815& Skype Chat& 53,996\\
		\hline
		Tor-Youtube& 54,294& VPN-Vimeo& 215,102\\
		\hline
	\end{tabular}\label{tb options}
\end{table}



We conduct our experiments on a workstation with an Intel(R) Core(TM) i9-9900K CPU$@$3.60GHz, 64.0 GB RAM$@$2133 MHz, 2 TB HD and two NVIDIA Corporation GP102 [TITAN X] GPUs. Each edge server ia simuylated with a TITAN X GPU running on Ubuntu 16.04, Python 3.6, CUDA 10.0 and PyTorch 1.3.1. The training process between edge servers is simulated using the PySyft framework \cite{ryffel2018generic}. We combine  the packet of all selected traffic types into one trace and equally divide the data into six local datasets, each being assigned to a discriminator (an edge server).
The network architecture and parameter setting for the generator, discriminator, and classifier are summarized in Table \ref{tb network}.

\begin{table}[htbp]\scriptsize
	\centering
	\caption{Simulation Setup}
	\begin{tabular}{|c|c|c|c|}
		\hline
		\multirow{2}{*}{Architecture Setup}&\multicolumn{3}{c|}{ Network Type}\\
		\cline{2-4}
		~&  Generator &  Discriminator&  Classifier\\
		\hline
		Input Size& 100$\times$1& 2500$\times$1& 2500$\times$1 \\
		\hline
		Output Size& 2500$\times$1& 2$\times$1 &$\#$ of Generator\\
		\hline
		Activation Function &LeakyReLU& Sigmoid&LeakyReLU\\
		\hline
		$\#$ of Layers& 8& \multicolumn{2}{c|}{ 6}\\
		\hline
		Optimizer& \multicolumn{3}{c|}{Adam with $\beta_1=0.5$ and $\beta_2=0.9$}\\
		\hline		
	\end{tabular}\label{tb network}
\end{table}


\begin{table*}[htbp]\scriptsize
	\centering
	\caption{Clustering Performance of FS-GAN and State-of-the-art Schemes}
	\begin{tabular}{|c|c|c|c|c|c|}
		\hline
		Scheme  & Training Loss& Training Method&RI& NMI& ACC\\
		\hline
		\hline
		\multirow{1}{*}{K-means++}& Mean Square Error &  Iterative Method&  0.71&0.30&0.32\\
		\hline
		\multirow{2}{*}{DEC}&Reconstruction and & Pretraining and&  \multirow{2}{*}{0.82}& \multirow{2}{*}{0.39}& \multirow{2}{*}{0.39}\\
		~&Cluster Assignment&Fine-tuning &  ~&~&\\
		\hline
		\multirow{2}{*}{DCN}&Reconstruction and & Pretraining and&  \multirow{2}{*}{0.81}& \multirow{2}{*}{0.31}& \multirow{2}{*}{0.35}\\
		~&K-means Loss&Joint Training &  ~&~&\\
		\hline
		\multirow{2}{*}{IDEC}&Reconstruction and & Pretraining and&  \multirow{2}{*}{0.85}& \multirow{2}{*}{0.42}& \multirow{2}{*}{0.40}\\
		~&Cluster Assignment&Joint Training &  ~&~&\\
		\hline
		\multirow{2}{*}{FS-GAN}&Adversarial and & Jointly Adversarial&  \multirow{2}{*}{\textbf{0.89}}& \multirow{2}{*}{\textbf{0.62}}& \multirow{2}{*}{\textbf{0.58}}\\
		~&Classification Loss& Training &  ~&~&\\
		\hline
		\end{tabular}\label{tb clusteringcomp}
	\end{table*}

\begin{table*}[htbp]\scriptsize
	\centering
	\caption{Model Training Time and Clustering Performance under Different Schemes}
	\begin{tabular}{|c c c c| c c| c c| c c|c c|}
		\hline
		\multirow{3}{*}{$N$}&\multirow{3}{*}{$n$}&\multirow{3}{*}{$E$}& \multirow{3}{*}{$B$}&\multicolumn{2}{c|}{Time}&\multicolumn{2}{c|}{\multirow{2}{*}{RI}}&\multicolumn{2}{c|}{\multirow{2}{*}{NMI}}&\multicolumn{2}{c|}{\multirow{2}{*}{ACC}}\\
		~&~&~&~&\multicolumn{2}{c|}{per-round (sec)}&&&&&&\\
		~&~&&&C-I&C-II&C-I&C-II&C-I&C-II&C-I&C-II\\
		\hline
		\hline		
		1 &12,000 &4&100&\multicolumn{2}{c|}{31.19} &\multicolumn{2}{c|}{0.91} &\multicolumn{2}{c|}{0.64}&\multicolumn{2}{c|}{0.63}\\
		\cline{1-12}
		6 &2,000 &4&100&34.32&39.16 &{0.89}&{0.89} &{0.62}&{0.43}&0.58&0.57\\
		\hline		
		\hline		
		3 &2,000 &4&100&34.72&35.43 &0.89&0.88 &0.61&0.42&0.60&0.58\\	
		\cline{1-12}
		6 &2,000 &4&100&34.32&39.16 &{0.89}&{0.89} &{0.62}&{0.43}&0.58&0.57\\
		\cline{1-12}
		9 &2,000 &4&100&38.03&37.21 &0.89&0.88 &0.60&0.40&0.63&0.57\\
		\hline
		\hline
		6 &500  & 4&100& 10.14&~11.36 &0.89&0.88&0.62&0.41&0.60&0.51\\
		\cline{1-12}
		6&1000 &4&100&18.33&18.14 &0.89&0.89 &0.60&0.47&0.56&0.55\\
		\cline{1-12}
		6 &2000 &4&100&34.32&39.16 &0.89&0.89 &0.62&0.43&0.58&0.57\\
		\hline
		\hline
		6 &2000  & 2&100&20.32&19.73 &0.89&0.88 &0.62&0.43&0.60&0.53\\
		\cline{1-12}
		6 &2000 &4&100&34.32&39.16 &0.89&0.89 &0.62&0.43&0.58&0.57\\
		\cline{1-12}
		6 &2000 &8&100&68.15&75.32 &{0.89}&{0.90} &{0.64}&{0.47}&0.61&0.54\\
		\hline
		\hline
		6 &2000  & 4&50&61.17&62.73 &{0.90}&{0.89} &{0.63}&{0.46}&0.57&0.60\\
		\cline{1-12}
		6 &2000 &4&100&34.32&39.16 &0.89&0.89 &0.62&0.43&0.58&0.57\\
		\cline{1-12}
		6 &2000 &4&200&21.50&24.13 &0.89&0.89 &0.57&0.43&0.57&0.57\\
		\hline
	\end{tabular}\label{tb FL}
\end{table*}

\begin{figure*}[ht]
\centering
\includegraphics[width=16cm]{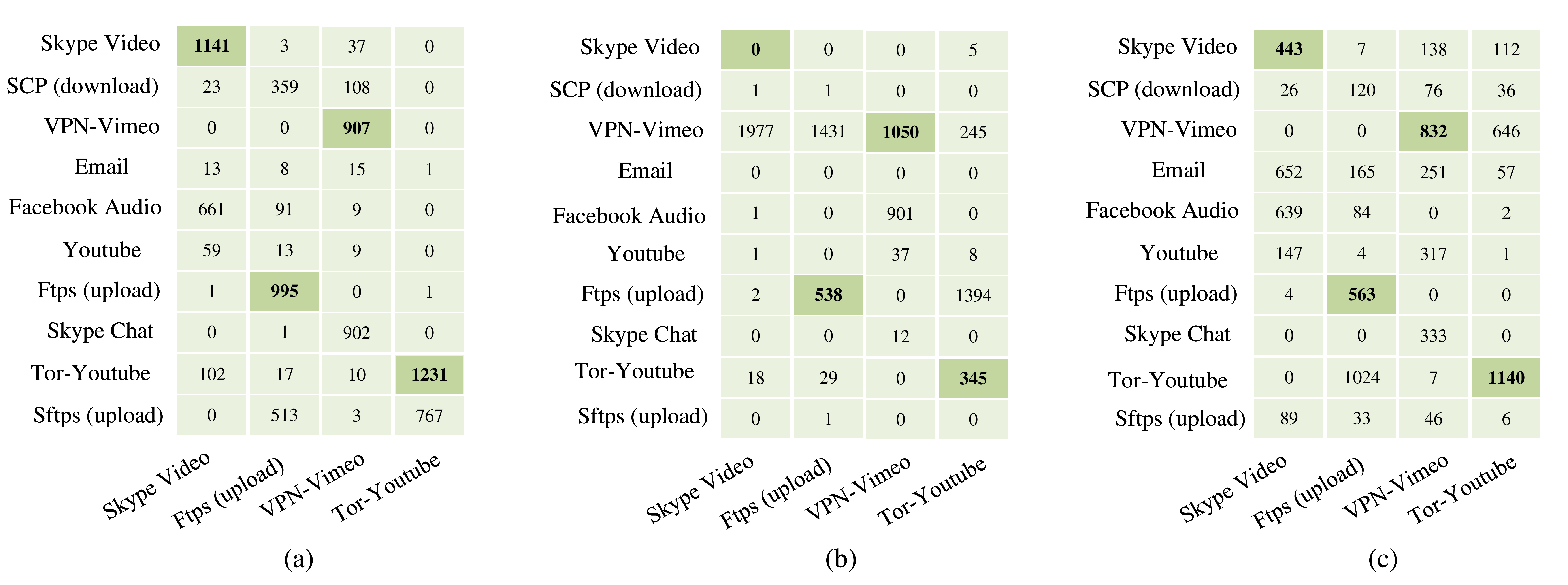}
\caption{\footnotesize{Clustering results of 10 service traffic data using: (a) FS-GAN, (b) K-means++, and (c) IDEC.}}
\label{fig matrix_new}
\end{figure*}

\begin{figure}
\centering
		\begin{minipage}[t]{0.45\linewidth}
			\center
			\includegraphics[width=4cm]{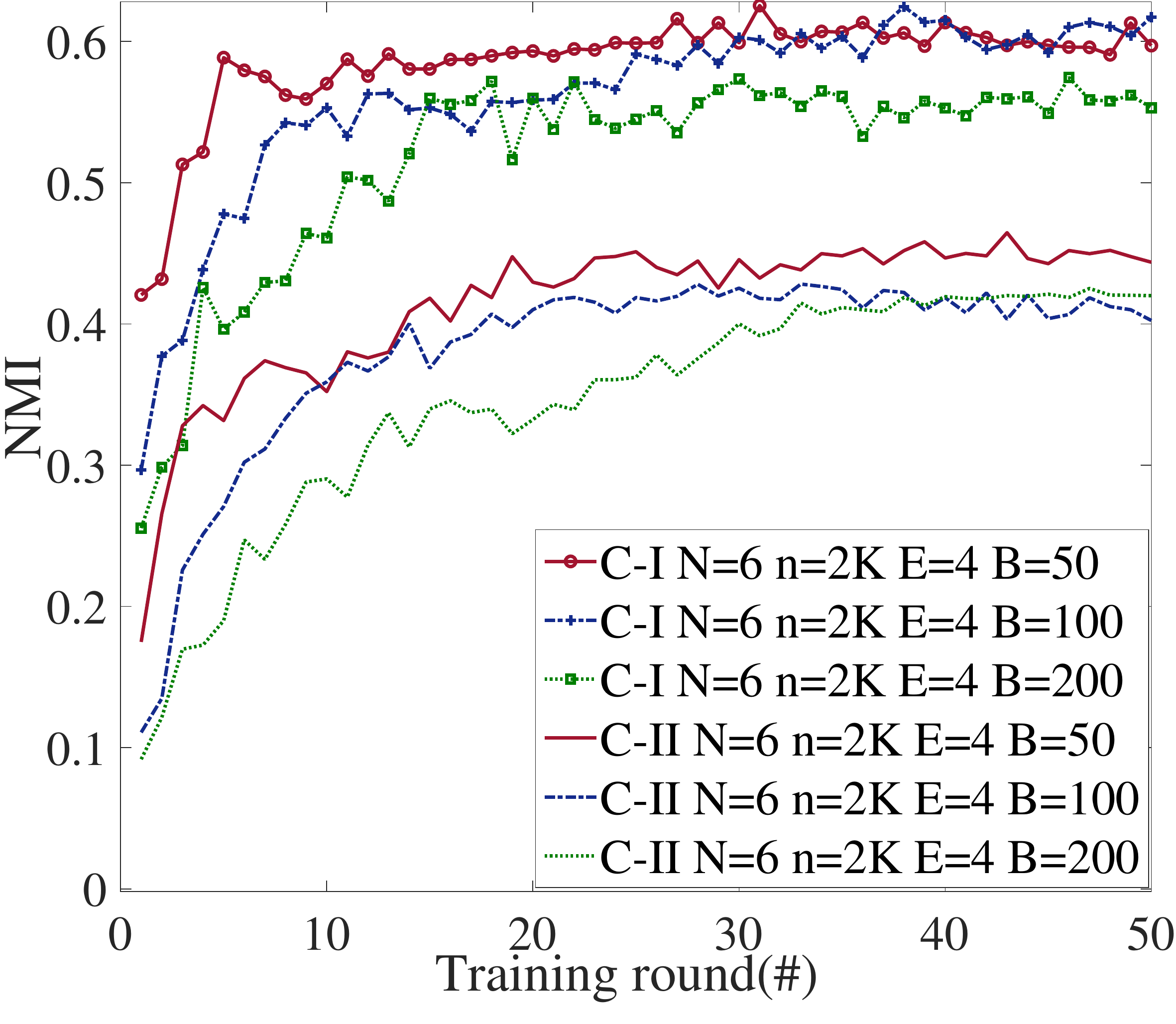}		
			\caption*{(a)}
		\end{minipage}
		\begin{minipage}[t]{0.45\linewidth}
			\center
			\includegraphics[width=4cm]{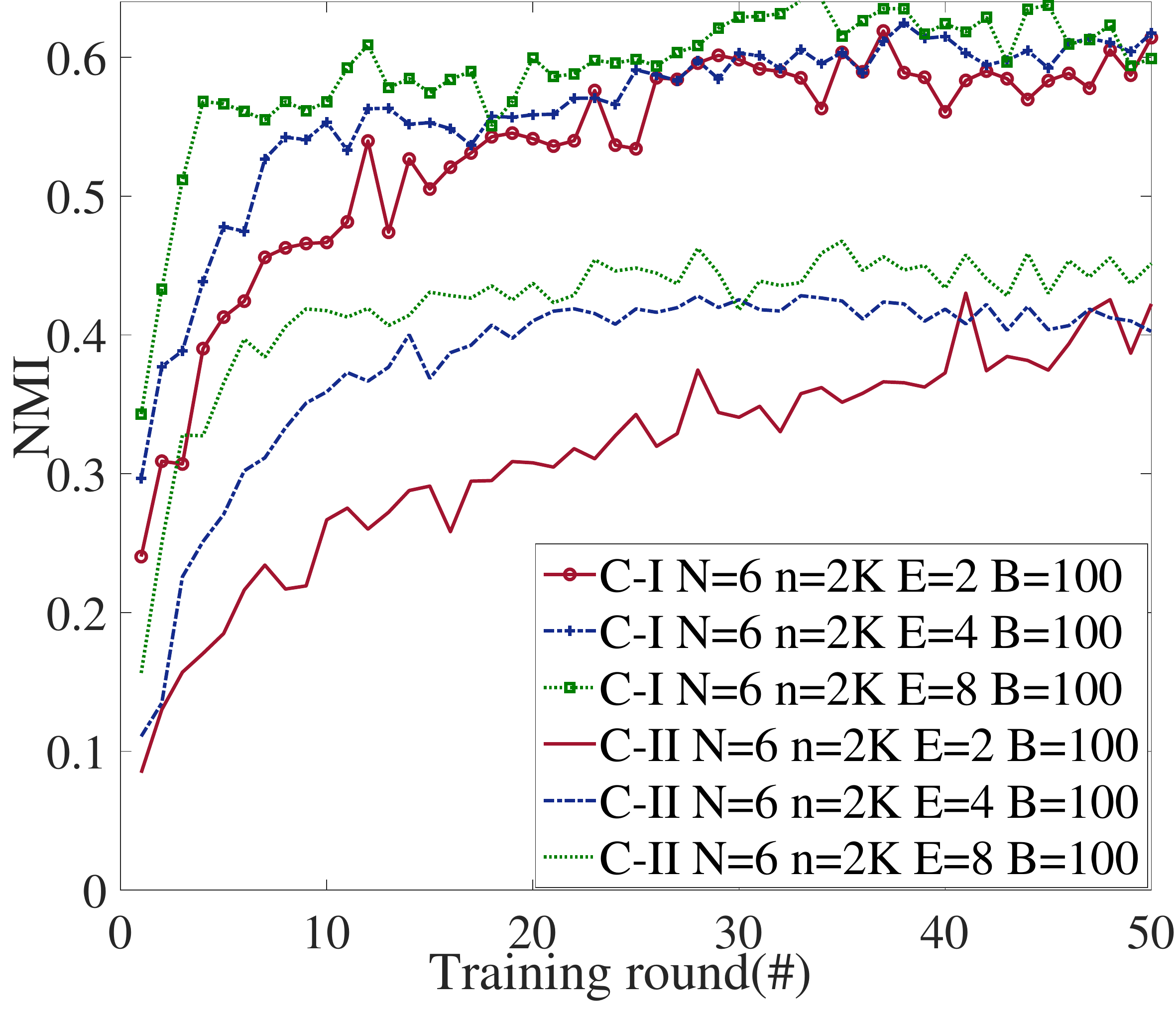}		
			\caption*{(b)}
		\end{minipage}
		\begin{minipage}[t]{0.45\linewidth}
			\center
			\includegraphics[width=4cm]{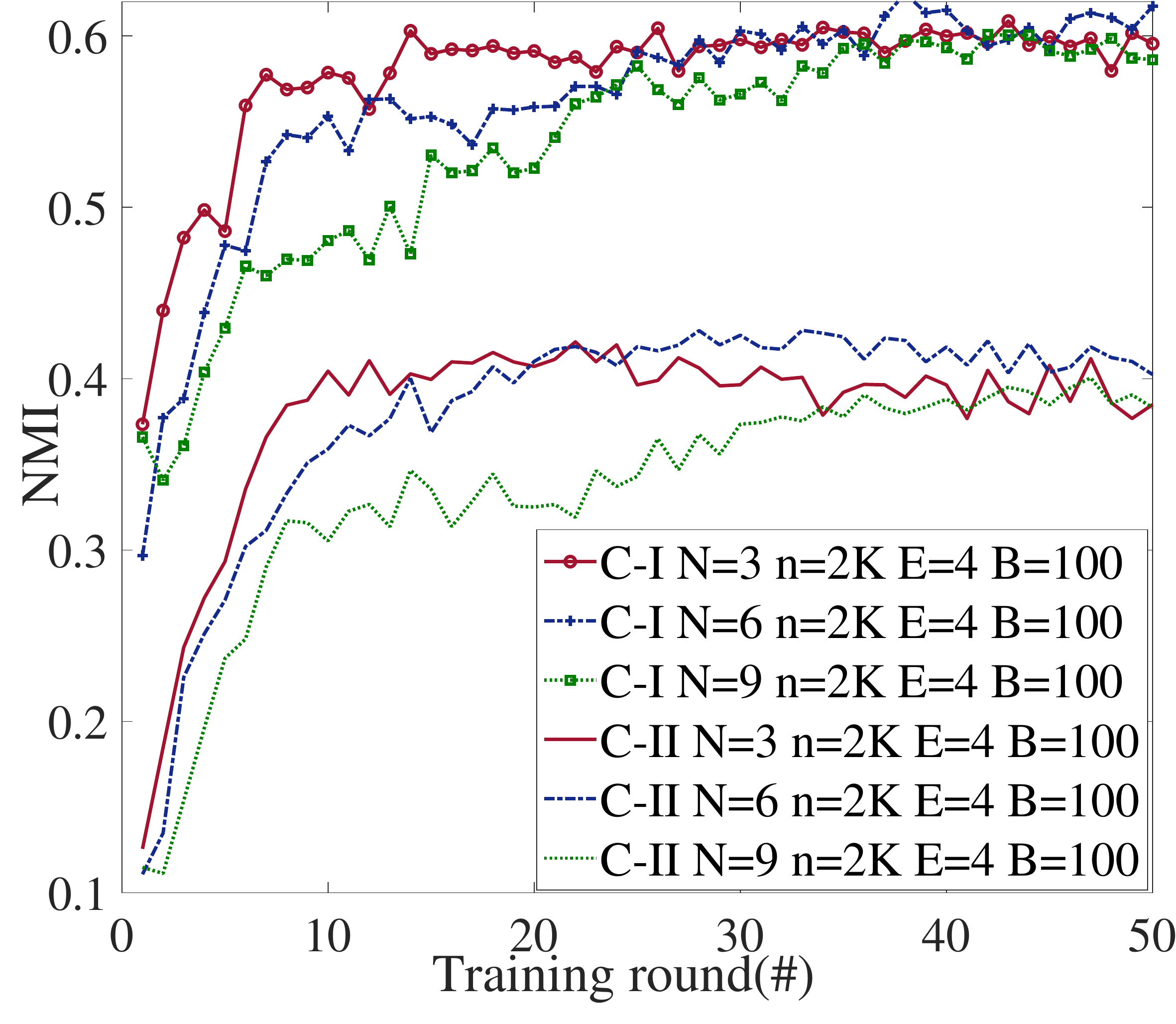}		
			\caption*{(c)}
		\end{minipage}
		\begin{minipage}[t]{0.45\linewidth}
			\center
			\includegraphics[width=4cm]{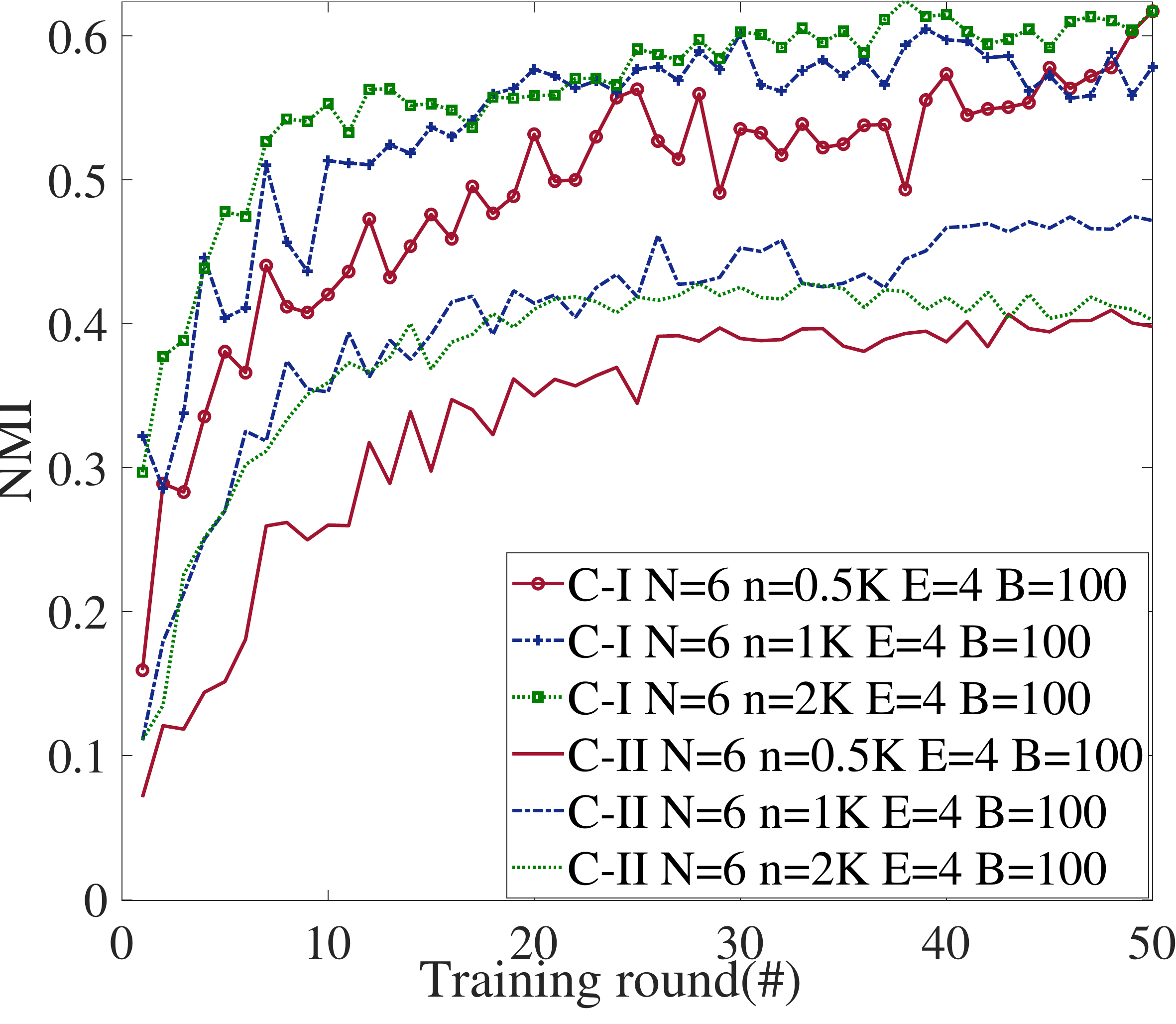}		
			\caption*{(d)}
		\end{minipage}
	\caption{\footnotesize{Clustering performance of FS-GAN-\uppercase\expandafter{\romannumeral1} and FS-GAN-\uppercase\expandafter{\romannumeral2} based on real-world traffic datasets.}}
\label{fig FL_A}
\end{figure}

\begin{figure}[htbp]
	{
		\centering
		\begin{minipage}[t]{0.45\linewidth}
			\center
			\includegraphics[width=4cm]{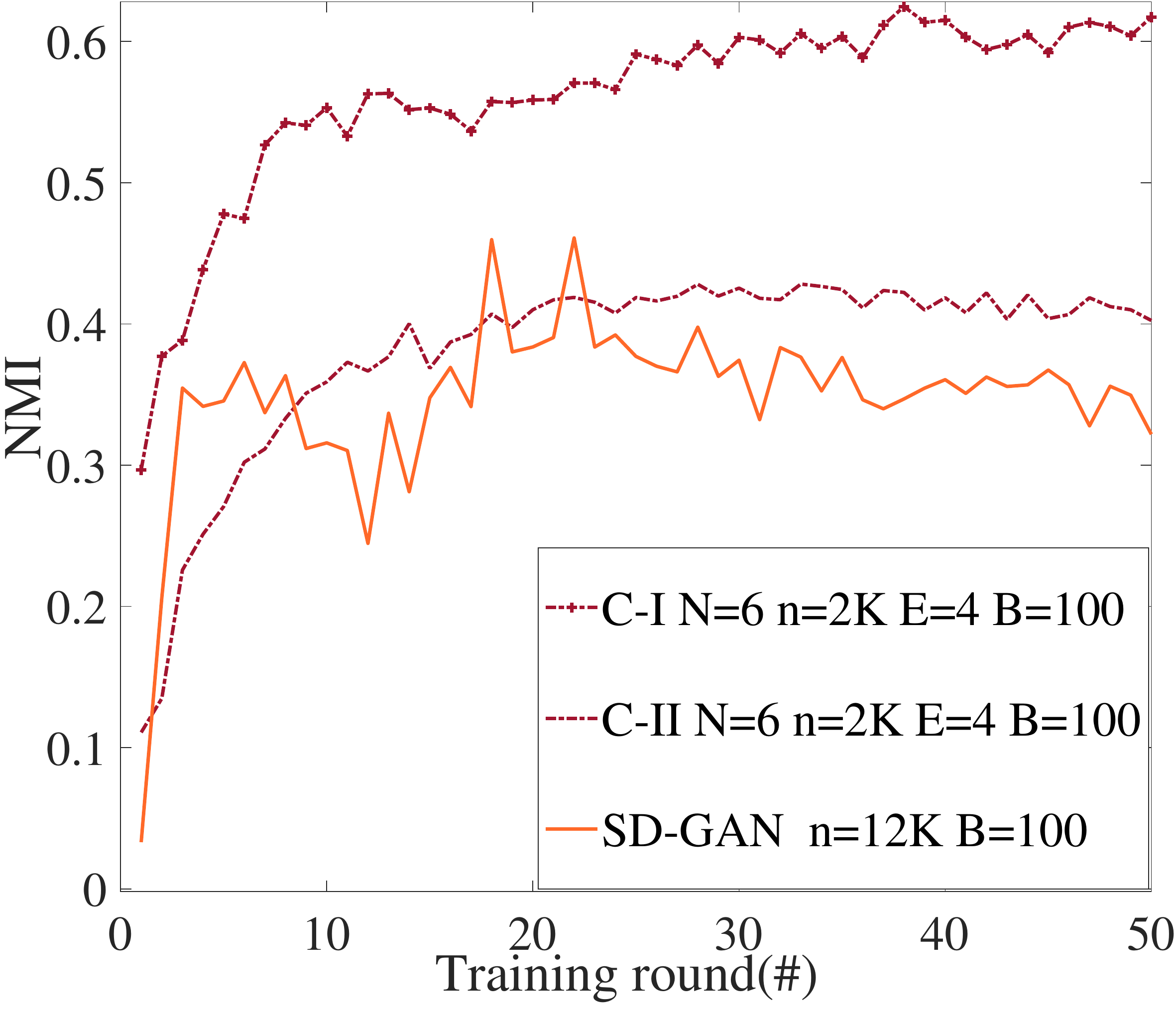}		
			\caption*{(a)}
		\end{minipage}
		\begin{minipage}[t]{0.45\linewidth}
			\center
			\includegraphics[width=4cm]{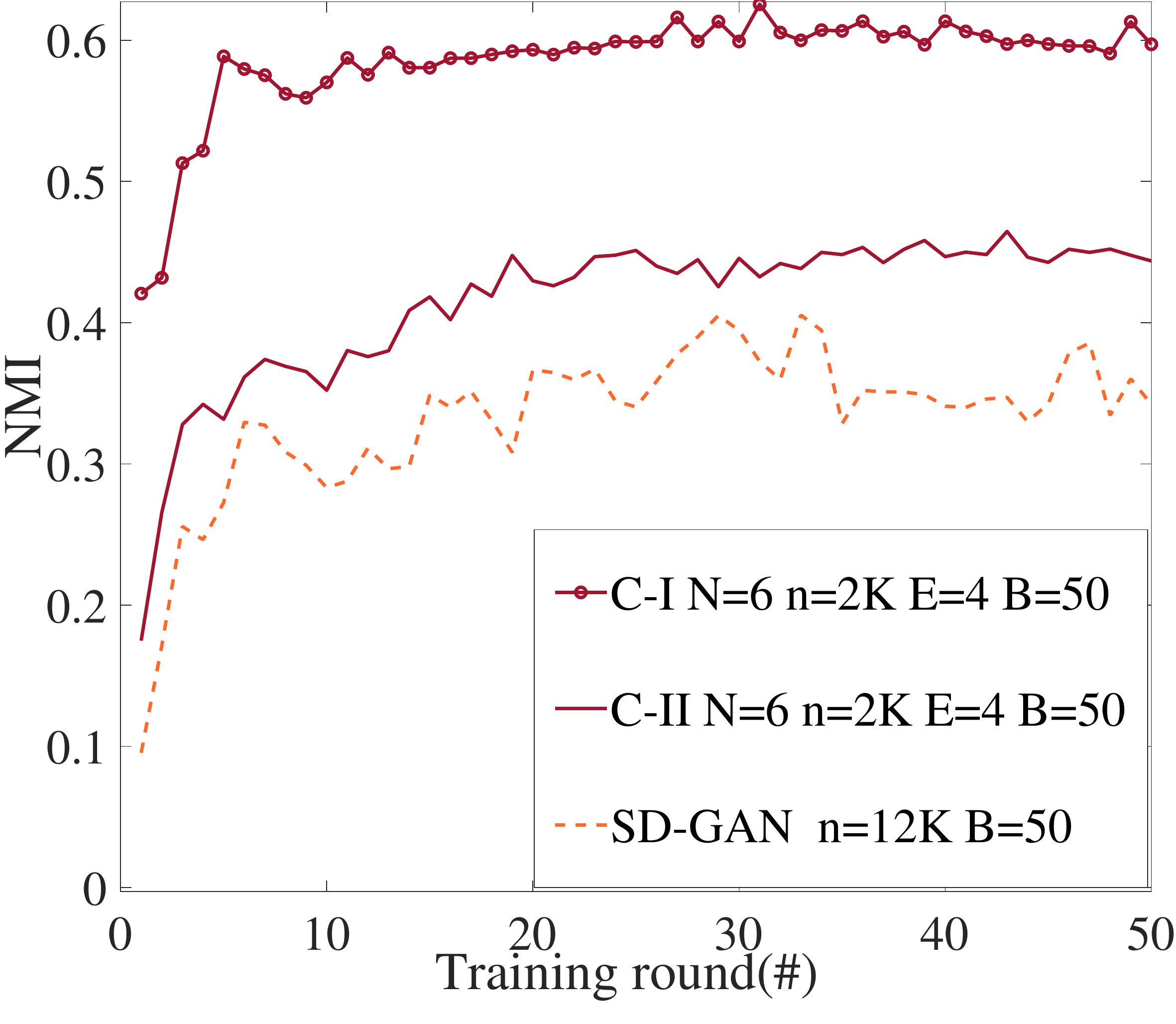}		
			\caption*{(b)}
		\end{minipage}
	}
	\centering
	\caption{\footnotesize{Clustering performance of FS-GAN and SD-GAN based on real-world traffic datasets.}}
	\label{fig SS-GAN}
\end{figure}

\subsection{Classification Performance }
As mention earlier, compared to existing clustering solutions, FS-GAN introduces a novel clustering solution for model training by autonomously creating synthetic data with pseudo-labels. In this subsection, we compare the clustering performance of FS-GAN with existing clustering solutions. We consider three commonly adopted performance metrics:

\noindent {\bf (1) Rand Index (\mbox{RI}):} measures the fraction of samples pairs being correctly clustered, including similar samples being classified into the same category and different samples being classified into different categories. Formally, \mbox{RI} is defined as follows:
\begin{equation}\label{metric RI}
	\begin{aligned}
	\mbox{RI}=\frac{\mbox{TP}+\mbox{TN}}{\mbox{TP}+\mbox{TN}+\mbox{FN}+\mbox{FP}},	
	\end{aligned}
\end{equation}
where $\mbox{TP}$ is the number of two same-type packets being assigned to the same traffic. $\mbox{TN}$ refers to the number of different-type packet pairs being assigned to different service types. $\mbox{FN}$ is the number of  same-type packets pairs being assigned to different service types. Finally, $\mbox{FP}$ is the number of different-type packet pairs being assigned to one service type. RI evaluates the clustering performance by judging whether samples of the same type fall into the same category. In particular, the value of RI must be grater than or equal to 0 and it approaches 1 if the values of $\mbox{FN}$ and $\mbox{FP}$ approach zero.

\noindent {\bf (2) Normalized Mutual Information (NMI):} measures the uncertainty that can be reduced about the ground truth clustering result when the clustering solution is given. Let $U$ be the ground truth class and $V$ be the resulting label of the proposed clustering algorithm. NMI is defined as:
\begin{equation}\label{metric NMI}
\begin{aligned}
\mbox{NMI}(U,V)=\dfrac{2 \mbox{MI}(U,V)}{H(U)+H(V)},
\end{aligned}
\end{equation}
where $\mbox{MI}(U,V)$  is the mutual information between $U$ and $V$, and $H(*)$ is the entropy. We have  

\noindent {\bf (3) Unsupervised Accuracy (\mbox{ACC}):} measures the best one-to-one relationship between the resulting label of the proposed clustering solution and the ground-truth classes. \mbox{ACC} is defined as:
\begin{equation}\label{metric ACC}
\begin{aligned}
\mbox{ACC}={1 \over n}\max_\sigma {\sum_{i=1}^n \mathrm{1}_{y_i=\sigma(x_i)}}
\end{aligned}
\end{equation}
where $x_i$ and $y_i$ refer respectively, to the labels of the clustering solution and the ground-truth class for each data sample $i$. $\sigma$ represents all the possible one-to-one permutations between any two class labels.

Note that each of the above three metrics takes values between $0$ and $1$. The larger the better is the clustering performance.

In Table \ref{tb clusteringcomp}, we compare FS-GAN with four clustering solutions: K-means++, DEC\cite{xie2016DEC}, IDEC\cite{guo2017IDEC}, and DCN\cite{yang2017DCN}. It can be observed that FS-GAN achieves the best performance of all tested solutions in all three performance metrics. In particular, DEC, IDEC, and DCN use deep neural networks to recover different latent representations for clustering. Compared to K-means++, which directly searches for cluster centers or centroids of the data samples, deep-learning-based solutions often achieve better performance. 
FS-GAN takes a novel approach by first generating synthetic data samples with pseudo-labels and then training a classifier based on the labeled dataset. In some senses, this approach addresses the poor performance caused by the lack of labelled dataset and can result in improved performance when the quality of the  pseudo-labeled synthetic data samples is high.

\blu{To compare FS-GAN with the most state-of-the-art self-supervised learning solution, we consider a recent extension of the DEC algorithm, called semi-supervised DEC (SDEC)\cite{REN2019121}, which incorporates the semi-supervised information learned from the labelled data samples in DEC to further improve the clustering performance. We use the SDEC as the pretext tasks and then include the peudo-labelled data samples into the self-supervised GAN algorithm to train the model, we referred to this approach as the SD-GAN. Note that, compared to FS-GAN, SD-GAN requires labelled datasets to learn a prior information to guide the learning process, it also requires an extra neural network model to perform the data clustering and peudo-labelling. In Fig. \ref{fig SS-GAN}, we compare FS-GAN with SD-GAN with different setups under different number of training rounds, we can observe that FS-GAN-I outperforms SD-GAN and FS-GAN-II achieves similar performance in terms of NMI. } 


To compare the impact of  model training parameters on the computational loads of FS-GAN under different setups,
we present in Table \ref{tb FL} the running time per round of computation for FS-GAN in a fixed hardware and software environment under various performance metrics for 50 rounds of model training.
We mainly evaluate the impact of 
four key parameters, including the number  of local datasets, size  of each dataset, the number of local training steps ($N$, $n$, $E$) between two consequent global model coordination, and the mini-batch size $B$.
The learning results under one centralized dataset is also presented. We can see that both the learning time and  performance of FS-GAN are slightly degraded compared with a centralized setting, because the model aggregation procedure is no longer needed.
It can be observed that the computational load is most sensitive to the size of the dataset. Compared to scheme C-I, Scheme C-II always results in slightly more computational load for training. This is because, in scheme C-I, both generators and discriminators are coordinated, which accelerates convergence of the model training and leads to reduced overall computational time. The improved performance offered by Scheme C-I over C-II can be further shown in the  \mbox{NMI} and \mbox{ACC} metrics. More specifically, C-I always exhibits a more accurate clustering performance than C-II when the number of training rounds is fixed.
C-I and C-II offer similar performance in terms of RI. This is because RI measures the combined performance of the correct solutions over all the clustering results, instead of the highest performance among individual classes  as \mbox{NMI} and \mbox{ACC}. Therefore, the performance difference in \mbox{RI} achieved by different schemes  is always smaller than that of the other two performance metrics. 

In Fig. \ref{fig matrix_new}, we present the clustering results when FS-GAN is used to classify the 10 services in Table \ref{tb options}. Once again, FS-GAN delivers superior performance to IDEC and K-means++. Note that since K-means++ directly separates data samples based on the centroid, it tends to cluster most of the samples into a single or a limited number of clusters.

\begin{table}[htbp]\scriptsize
	\centering
	\caption{Performance under Different Classifier Level $\lambda$}
	\begin{tabular}{ |c|c| c| c|c|}
		\hline
		\multicolumn{2}{|c|}{{Diversity Parameter}}& $\lambda=0.1$& $\lambda=0.5$& $\lambda=5$\\
		\hline
		\multirow{3}{*}{{FS-GAN (C-I)}}& RI& 0.89& 0.89 &\textbf{0.90}\\
		\cline{2-5}
		~& NMI& 0.60&0.60&\textbf{0.64}\\
		\cline{2-5}
		~& ACC& 0.57&0.58&\textbf{0.60}\\
		\hline
		\multirow{3}{*}{{FS-GAN (C-II)}}& RI& 0.88& \textbf{0.89}&0.88\\
		\cline{2-5}
		~& NMI& 0.43& \textbf{0.47}&0.40\\
		\cline{2-5}
		~& ACC& 0.52& \textbf{0.57}&0.51\\
		\hline
		\end{tabular}\label{tb arg_classifier}
\end{table}

\begin{figure}[ht]
	\centering
	\includegraphics[width=9cm]{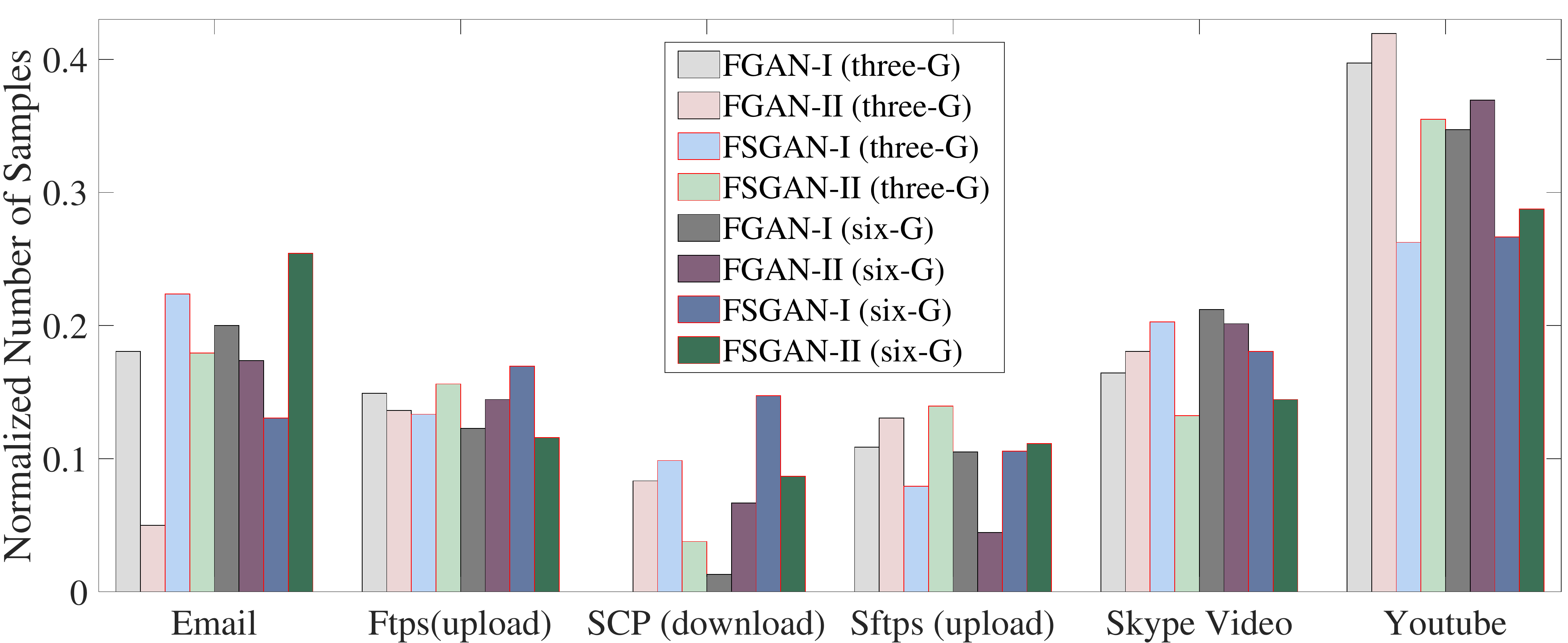}
	\centering
	\caption{\footnotesize{Distribution of the assigned labels under different schemes.}}\label{fig hist}
\end{figure}

\begin{table*}[htbp]\scriptsize
	\centering
	\caption{Quality of Synthesized Under Different Scenarios}
	\begin{tabular}{|c|c|c|c|c|c|c|c|c|}
		\hline
		\multirow{2}{*}{Metric}&\multicolumn{2}{c|}{\textbf{F-GAN(C-I)}}&\multicolumn{2}{c|}{\textbf{F-GAN(C-II)}}&\multicolumn{2}{c|}{\textbf{FS-GAN(C-I)}}&\multicolumn{2}{c|}{\textbf{FS-GAN(C-II)}}\\
		\cline{2-9}
		~&Three-G & Six-G& Three-G & Six-G & Three-G & Six-G & Three-G & Six-G \\
		\hline
		\hline
		KL Divergence& 5.643& 0.360&0.226&0.224&0.091&\textbf{0.043}&0.188& 0.099\\
		\hline
		JS Divergence& 0.086 &0.062&0.054&0.051&0.022&\textbf{0.011}& 0.041& 0.025\\
		\hline
		Wasserstein Distance&0.647& 0.619 &1.013&0.578&0.353&\textbf{0.286}& 0.484& 0.387\\
		\hline
	\end{tabular}\label{tb JSD_G}
\end{table*}

In Fig. \ref{fig FL_A}, we compare the convergence performance based on \mbox{NMI} under different model parameters. We can observe that C-I always offers better convergence performance than C-II. This result is consistent with the previous observation. When the size of local dataset is the same, the convergence performance is closely correlated with this size. Specifically, the larger the dataset size, the faster is the convergence of the model training process. Note that the fluctuations in \mbox{NMI} is due to the interaction between generators and discriminators in the local data synthesis as well as the global model coordination.



As mentioned earlier, the classifier helps increasing the diversity of the synthesized samples produced by generators. At the same time, it interferes with the adversarial training process between the generator and discriminator.
We investigate the impact of the classifier on the data clustering performance by adjusting the diversity hyper-parameter $\lambda$ in Equation (\ref{eq localMixGAN}). Intuitively, when the penalty brought by the classifier becomes larger than that of the discriminator,  generators will be encouraged to produce more diversified samples even if these samples may not follow the same distribution as 
the real data.
In Table \ref{tb arg_classifier}, we present the clustering performance of FS-GAN for different values of $\lambda$. We can observe that a very large or small value of $\lambda$ will not typically offer the best performance 
The optimal $\lambda$ varies with the model setup. Finding the optimal $\lambda$ that achieves the right balance between classification and discrimination will be left for future work.

\subsection{Performance Evaluation of Traffic Data Synthesis }
As mentioned earlier, FS-GAN is more than just a traffic classification approach. It can also learn from the decentralized datasets and produce synthetic data samples that capture the distribution of real data associated with unknown services. In this subsection, we evaluate this aspect of FS-GAN.

We compare the data synthesis capability of FS-GAN to various extensions of multi-generator GANs with FL, which we refer to as F-GAN. In particular, we compare the following 4 different schemes:
\begin{itemize}
	\item[1)] F-GAN (three-G): 3 generators for each dataset without the classifier.
	\item[2)] FS-GAN (three-G): 3 generators for each dataset.
	\item[3)] F-GAN (six-G): 6 generators for each dataset without the classifier.
	\item[4)] FS-GAN (six-G): 6 generators for each dataset.
\end{itemize}

As mentioned earlier, one of the key challenges of GAN-based solutions is the possibility of triggering a model collapse problem where, in this case, different generators will only produce samples with limited variety regardless of the input. In Fig. \ref{fig hist}, we evaluate the diversity of synthesized data samples produced by different schemes. We compare the default numbers of synthetic samples produced by each scheme that are associated with different services when the same input is employed. It can be observed that different generators tend to produce different numbers of synthetic samples. For example, among all six types of services, SCP (download) is the least popular service to be synthesized with the least number of synthesized data samples, e.g., less than 0.02\% of total synthesized samples produced by F-GAN(C-I, six-G) falls into this service category.
We, however, observe that FS-GAN, especially C-I scheme with six generators, produces relatively uniform number of samples for different services. This means that FS-GAN offers the best performance in terms of balancing synthesized data samples.

To quantify the diversity of the synthesized samples, we use statistical distance metrics, which measure the difference between the distribution of the synthetic data samples and that of the real service traffic data. 
We consider three distance metrics: Kullback-Leibler (KL) divergence, Jensen Shannon (JS) divergence  \cite{JSD}, and Wasserstein (W) distance \cite{WD}, consider two random variables, $p$ and $q$, with respective distributions $p(x)$ and $q(x)$.
Table \ref{tb JSD_G} provides the distance results under both FS-GAN and F-GAN. FS-GAN achieves the smallest distance between synthetic and real data, where the JS divergence under FS-GAN-I is shown to be as low as 0.011, almost one fifth of the lowest distance result achieved by F-GAN. This means that the proposed FS-GAN is ideal to classify and synthesize  highly heterogeneous traffic flows with mixed services.

\section{Conclusions}
\label{Section_Conclusion}
In this paper, we proposed FS-GAN, a federated self-supervised learning framework to automatically recognize, classify, and synthesize different types of traffic  over a large number of decentralized datasets. FS-GAN consists of three components: local data synthesis, global model coordination, and self-supervised learning. We adopted an FL-like approach and proved that our jointly trained global model can simultaneously minimize the JSD between the distribution of real data across all the datasets and that of the synthesized data samples.It also maximizes the JSD among the distributions of data samples created by different generators. Simulation results show that FS-GAN can achieve significant performance improvement over state-of-art data clustering  solutions, and almost five times improvement over the federated GAN solutions in terms of data synthesis diversity.



\ifCLASSOPTIONcaptionsoff
  \newpage
\fi
\bibliography{fsganbib}
\bibliographystyle{IEEEtran}




\begin{IEEEbiography}[{\includegraphics[width=1.1in,height=1.3in,clip,keepaspectratio]{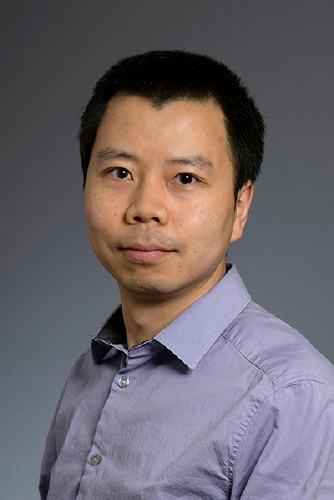}}]{Yong Xiao} (Senior Member, IEEE) received his B.S. degree in electrical engineering from China University of Geosciences, Wuhan, China in 2002, M.Sc. degree in telecommunication from Hong Kong University of Science and Technology in 2006, and his Ph. D degree in electrical and electronic engineering from Nanyang Technological University, Singapore in 2012. He is now a professor in the School of Electronic Information and Communications at the Huazhong University of Science and Technology (HUST), Wuhan, China. He is also with Peng Cheng Laboratory, Shenzhen, China and Pazhou Laboratory (Huangpu), Guangzhou, China. He is the associate group leader of the network intelligence group of IMT-2030 (6G promoting group) and the vice director of 5G Verticals Innovation Laboratory at HUST. 
His research interests include machine learning, game theory, distributed optimization, and their applications in semantic communications and semantic-aware networks, cloud/fog/mobile edge computing, green communication systems, wireless communication networks, and Internet-of-Things (IoT).
\end{IEEEbiography}

\begin{IEEEbiography}[{\includegraphics[width=1.1in,height=1.3in,clip,keepaspectratio]{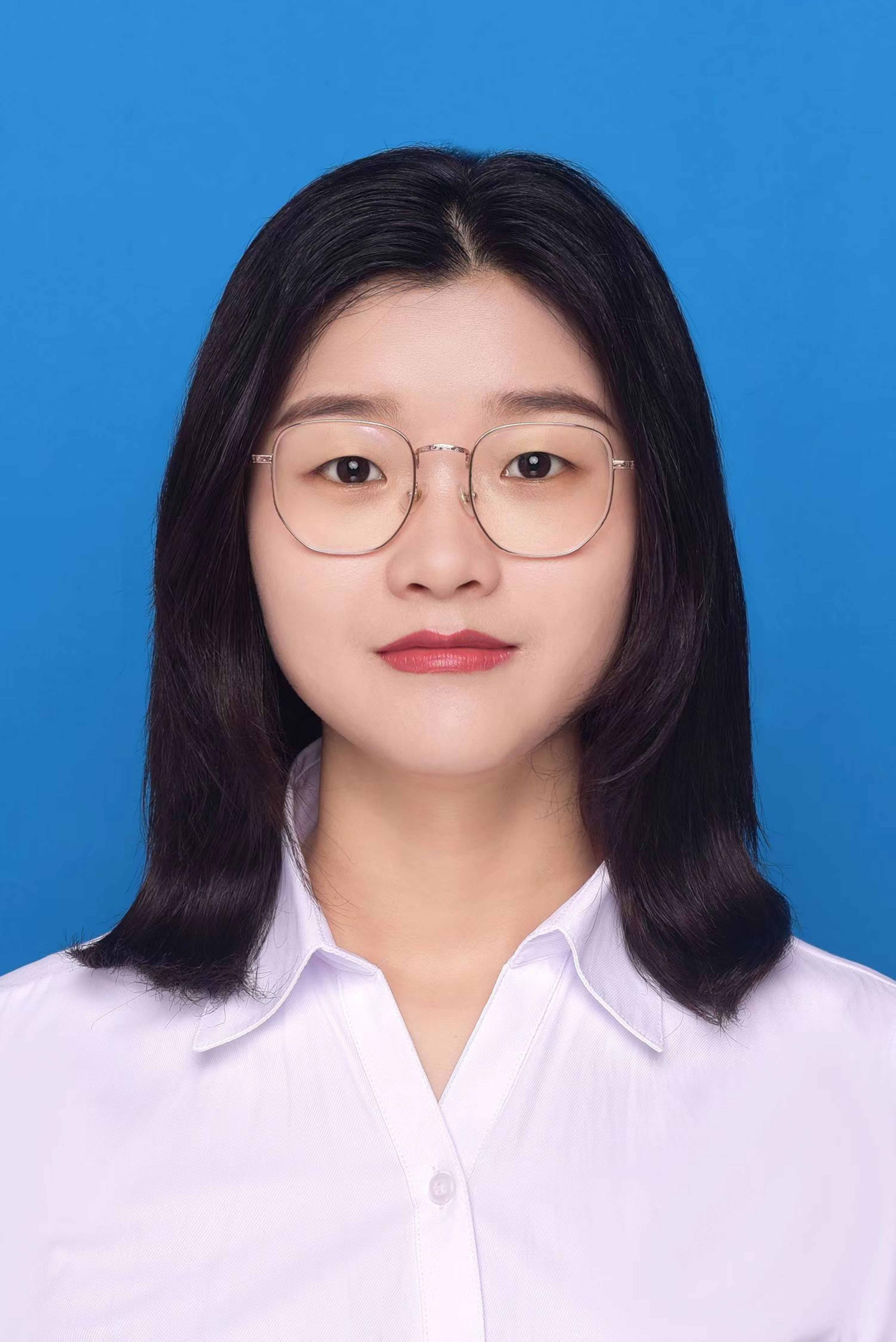}}]{Rong Xia} received her B.S. degree and M.Sc. degree both in information and communication engineering from Huazhong University of Science and Technology, Wuhan, China in 2018 and 2022, respectively. Her research interests include federated edge intelligence and edge computing networks.
\end{IEEEbiography}

\begin{IEEEbiography}[{\includegraphics[width=1.1in,height=1.3in,clip,keepaspectratio]{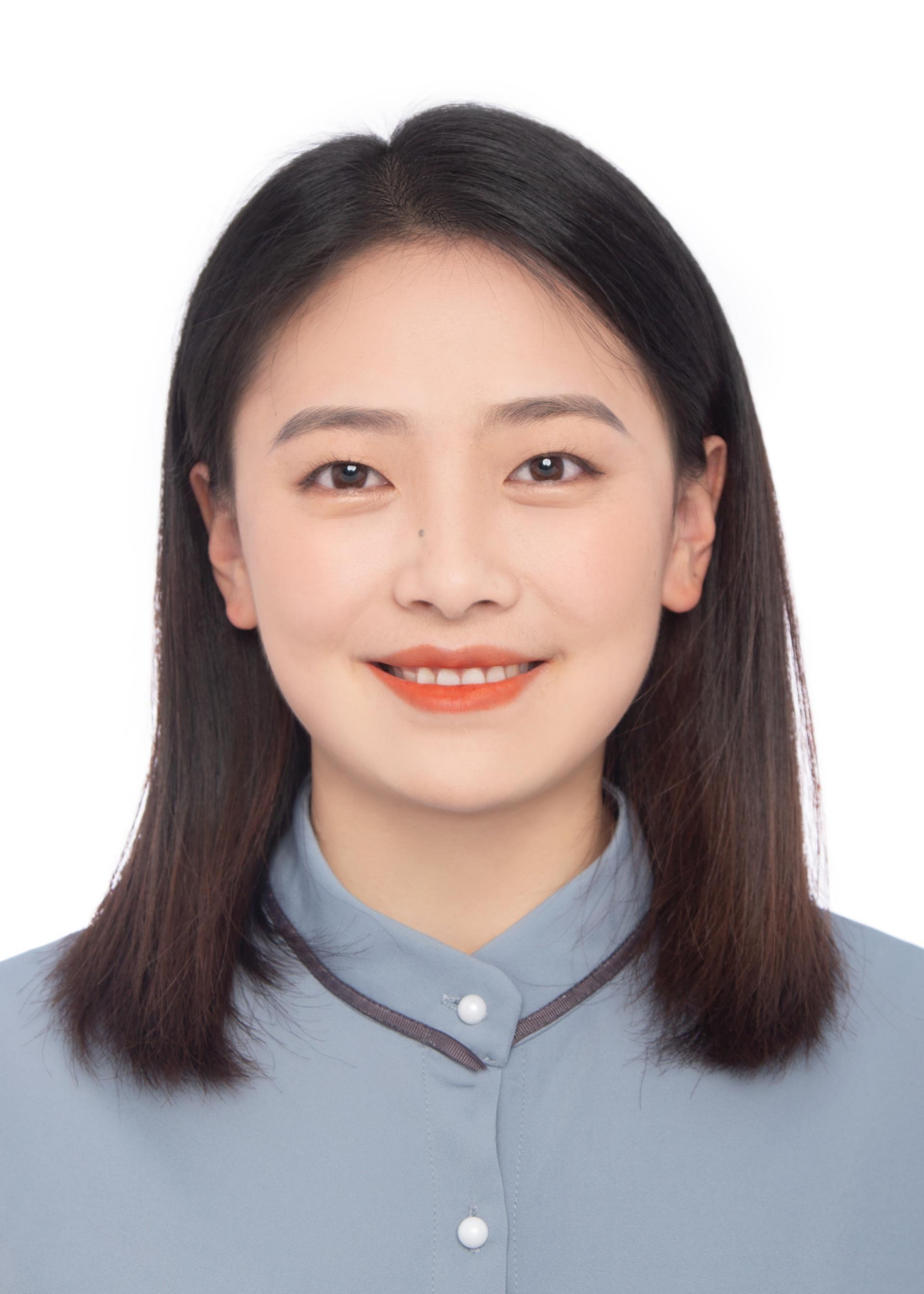}}]{Yingyu Li} (Member, IEEE) received the B.Eng. degree in electronic information engineering and the Ph.D. degree in circuits and systems from the Xidian University, Xi'an, China, in 2012 and 2018, respectively. From 2014 to 2016, she was a Research Scholar with the Department of Electronic Computer Engineering at the University of Houston, TX, USA. She was a postdoctoral researcher in the School of Electronic Information and Communications at Huazhong University of Science and Technology from 2018 to 2021. She is now an associate professor at the School of Mechanical Engineering and Electronic Information, China University of Geosciences (Wuhan). Her research interests include semantic communications, edge intelligence, green communication networks, and IoT.
%
\end{IEEEbiography}

\begin{IEEEbiography}[{\includegraphics[width=1.1in,height=1.3in,clip,keepaspectratio]{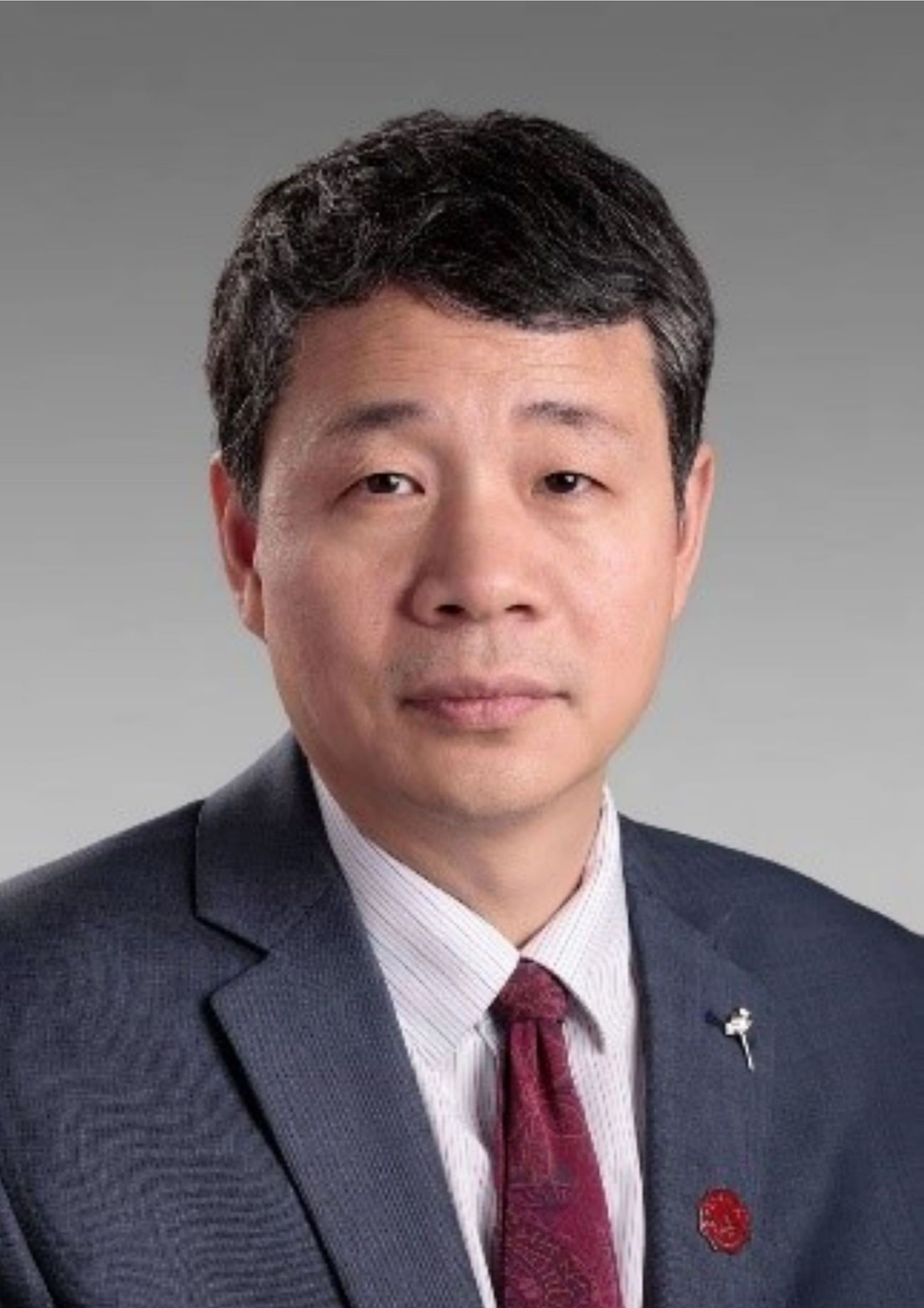}}]{Guangming Shi} (Fellow, IEEE) received the M.S. degree in computer control and the Ph.D. degree in electronic information technology from Xidian University, Xi’an, China, in 1988, and 2002, respectively. He was the vice president of Xidian University from 2018 to 2022. Currently, he is  the Vice Dean of Peng Cheng Laboratory and a Professor with the School of Artificial Intelligence, Xidian University. He is an IEEE Fellow, the chair of IEEE CASS Xi’an Chapter, senior member of ACM and CCF, Fellow of Chinese Institute of Electronics, and Fellow of IET. He was awarded Cheung Kong scholar Chair Professor by the ministry of education in 2012. He won the second prize of the National Natural Science Award in 2017. His research interests include Artificial Intelligence, Semantic Communications, and Human-Computer Interaction.
\end{IEEEbiography}

\begin{IEEEbiography}[{\includegraphics[width=1.1in,height=1.3in,clip,keepaspectratio]{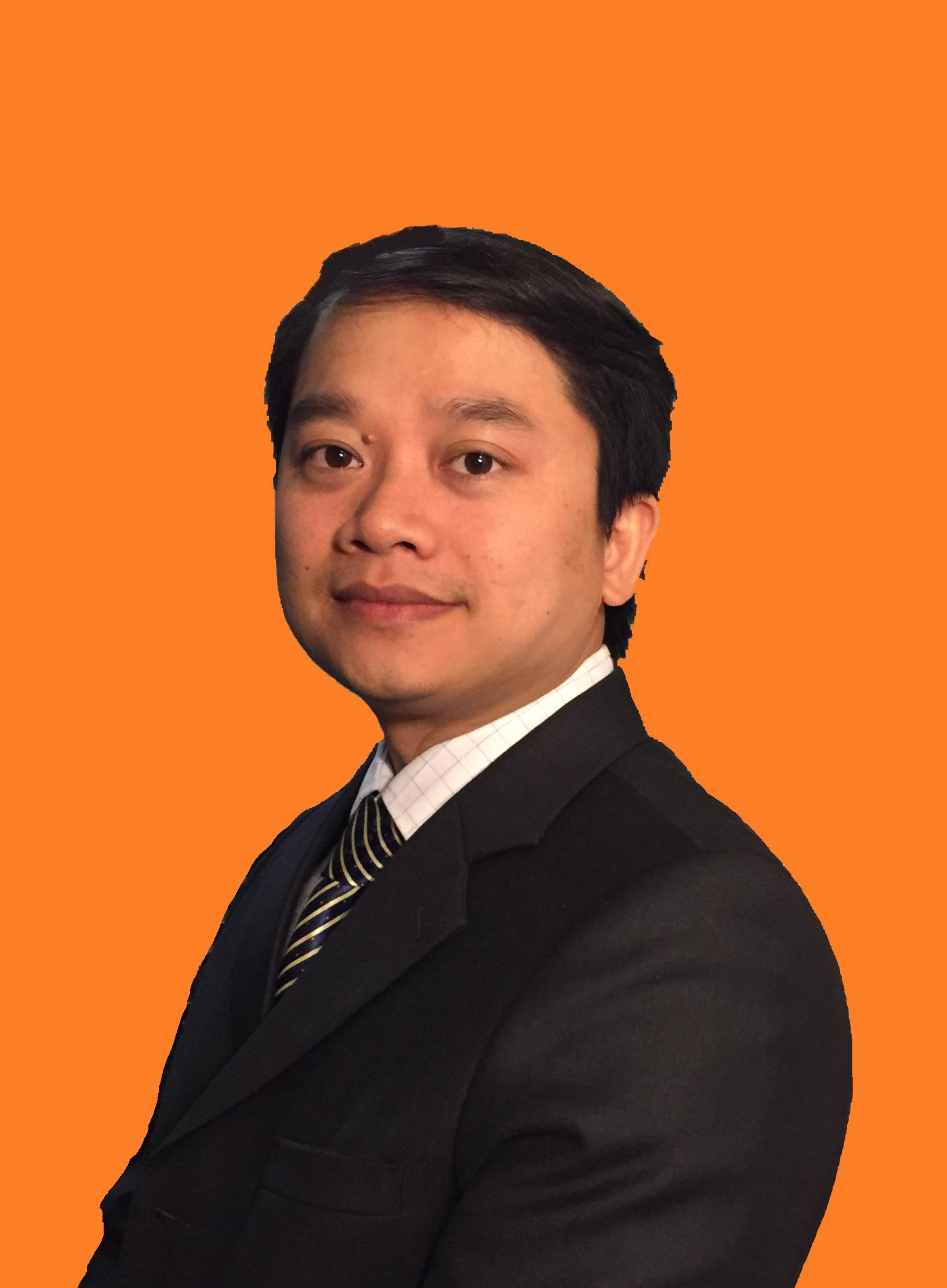}}]{Diep N. Nguyen} (Senior Member, IEEE) received the M.E. degree in electrical and computer engineering from the University of California at San Diego (UCSD), La Jolla, CA, USA, in 2008, and the Ph.D. degree in electrical and computer engineering from The University of Arizona (UA), Tucson, AZ, USA, in 2013. He is currently a Faculty Member with the Faculty of Engineering and Information Technology, University of Technology Sydney (UTS), Sydney, NSW, Australia. Before joining UTS, he was a DECRA Research Fellow with Macquarie University, Macquarie Park, NSW, Australia, and a Member of the Technical Staff with Broadcom Corporation, San Jose, CA, USA, and ARCON Corporation, Boston, MA, USA, and consulting the Federal Administration of Aviation, Washington, DC, USA, on turning detection of UAVs and aircraft, and the U.S. Air Force Research Laboratory, Wright-Patterson Air Force Base, OH, USA, on anti-jamming. His research interests include computer networking, wireless communications, and machine learning application, with emphasis on systems’ performance and security/privacy. Dr. Nguyen received several awards from LG Electronics, UCSD, UA, the U.S. National Science Foundation, and the Australian Research Council. He is currently an Editor, an Associate Editor of the IEEE Transactions on Mobile Computing, IEEE Communications Surveys \& Tutorials (COMST), IEEE Open Journal of the Communications Society, and Scientific Reports (Nature's).
\end{IEEEbiography}

\begin{IEEEbiography}[{\includegraphics[width=1.1in,height=1.3in,clip,keepaspectratio]{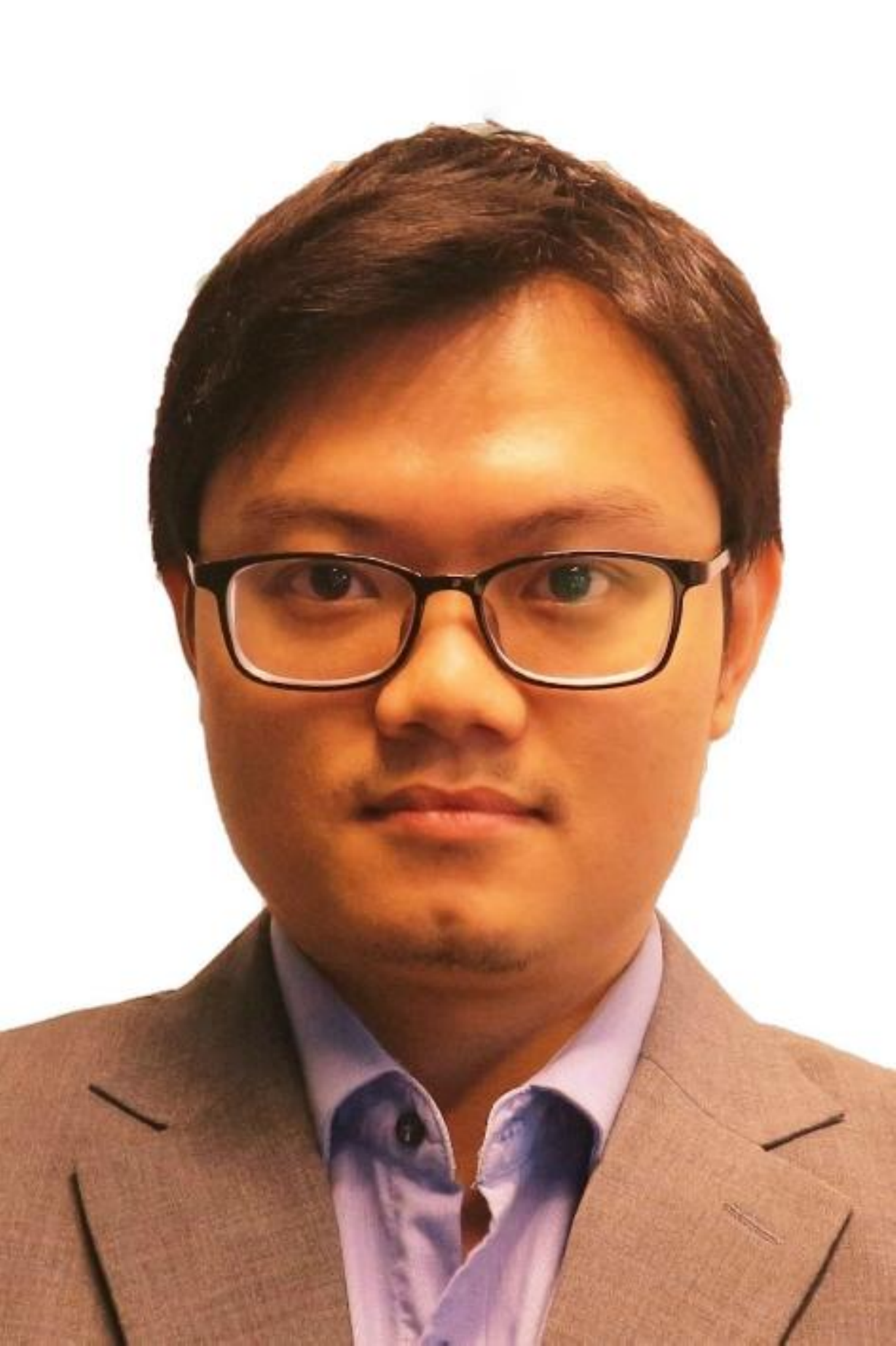}}]{Dinh Thai Hoang} (Senior Member, IEEE)  is currently a faculty member at the School of Electrical and Data Engineering, University of Technology Sydney, Australia. He received his Ph.D. in Computer Science and Engineering from the Nanyang Technological University, Singapore, in 2016. His research interests include emerging wireless communications and networking topics, especially machine learning applications in networking, edge computing, and cybersecurity. He has received several awards, including the Australian Research Council and IEEE TCSC Award for Excellence in Scalable Computing (Early Career Researcher). He is an Editor of IEEE Transactions on Wireless Communications, IEEE Transactions on Cognitive Communications and Networking, IEEE Transactions on Vehicular Technology, and Associate Editor of IEEE Communications Surveys \& Tutorials.
\end{IEEEbiography}

\begin{IEEEbiography}[{\includegraphics[width=1.1in,height=1.3in,clip,keepaspectratio]{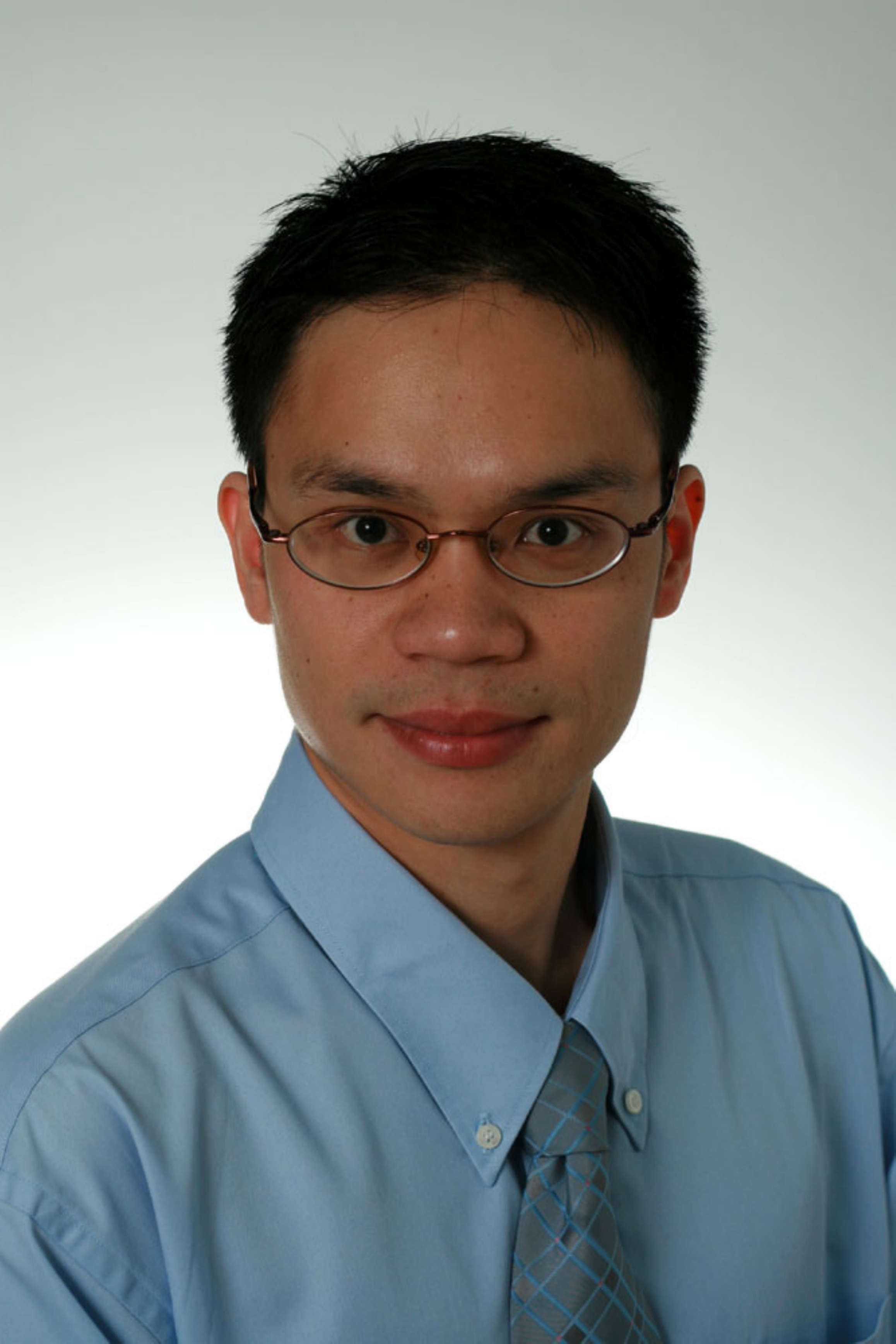}}]{Dusit Niyato} (Fellow, IEEE) is a professor in the School of Computer Science and Engineering, at Nanyang Technological University, Singapore. He received B.Eng. from King Mongkuts Institute of Technology Ladkrabang (KMITL), Thailand in 1999 and Ph.D. in Electrical and Computer Engineering from the University of Manitoba, Canada in 2008. His research interests are in the areas of Internet of Things (IoT), machine learning, and incentive mechanism design.
\end{IEEEbiography}

\begin{IEEEbiography}[{\includegraphics[width=1.1in,height=1.3in,clip,keepaspectratio]{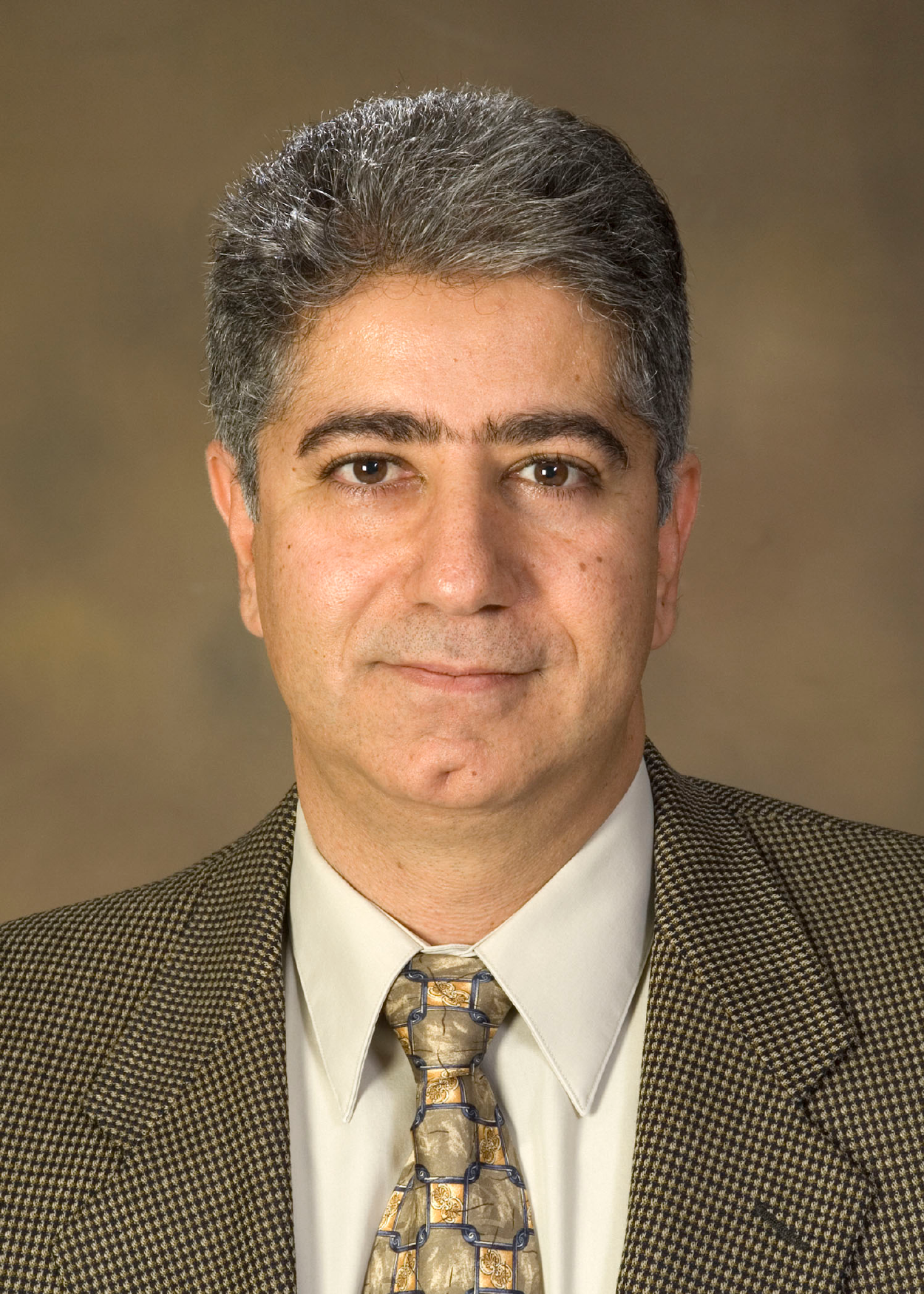}}]{Marwan Krunz}(Fellow, IEEE) is a Regents Professor at the University of Arizona. He holds the Kenneth VonBehren Endowed Professorship in ECE and is also a professor of computer science. He directs the Broadband Wireless Access and Applications Center (BWAC), a multi-university NSF/industry center that focuses on next-generation wireless technologies. He also holds a courtesy appointment as a professor at University Technology Sydney. Previously, he served as the site director for  Connection One, an NSF/industry-funded center of five universities and 20+ industry affiliates. Dr. Krunz’s research is in the fields of wireless communications, networking, and security, with recent focus on applying AI and machine learning techniques for protocol adaptation, resource management, and signal intelligence. He has published more than 320 journal articles and peer-reviewed conference papers, and is a named inventor on 12 patents. His latest h-index is 60. He is an IEEE Fellow, an Arizona Engineering Faculty Fellow, and an IEEE Communications Society Distinguished Lecturer (2013-2015). He received the NSF CAREER award. He served as the Editor-in-Chief for the IEEE Transactions on Mobile Computing. He also served as editor for numerous IEEE journals. He was the TPC chair for INFOCOM’04, SECON’05, WoWMoM’06, and Hot Interconnects 9. He was the general vice-chair for WiOpt 2016 and general co-chair for WiSec’12. Dr. Krunz served as chief scientist/technologist for two startup companies that focus on 5G and beyond wireless systems. 
\end{IEEEbiography}

\appendices
\section{Proof of Proposition 1}\label{apend1}
\begin{IEEEproof}
%
	According to the multi-player zero-sum game in Equation (\ref{eq localMixGAN}), classifier ${C_d}$ tries to maximize the softmax quantity $V(C)$, where
	\begin{equation}\label{eq V(C)}
	\begin{split}
	V(C)=&\int_{\bx} \sum_{m=1}^{M}\pi_{m}  P_{G^{m}_d}(\bx)\log C^{m}_d({\bx}) dx\\
	=&\int_{\bx} \pi_{1}  P_{G^{1}_d}(\bx)\log (1-\sum_{m=2}^{M}C^{m}_d({\bx}))dx\\
	&+\int_{\bx}\sum_{m=2}^{M}\pi_{m}  P_{G^{m}_d}(\bx)\log C^{m}_d({\bx}) dx.
	\end{split}
	\end{equation}
	
	In order to get the optimal classifier, we first calculate the functional derivative w.r.t. ${C^{m}_d}, m=2, 3, ..., M$, we get:
	\begin{equation}\label{eq V(C) derivative1}
	\begin{aligned}
	\frac{\partial V(C)}{\partial C^{m}_d(\bx)}&=\dfrac{\pi_mP_{G^{m}_d}(\bx)}{ C^{m}_d(\bx)}-\dfrac{\pi_1P_{G^{1}_d}(\bx)}{ C^{m}_d(\bx)}.
	\end{aligned}
	\end{equation}
	
	Then, by setting the above equation to zero for $m\in\{2, 3, ..., M\}$, we obtain:
	\begin{equation}\label{eq V(C) derivative2}
	\begin{aligned}
	\dfrac{\pi_1P_{G^{1}_d}(\bx)}{ C^{1}_d(\bx)}=\dfrac{\pi_m P_{G^{m}_d}(\bx)}{ C^{m}_d(\bx)}, \forall m\in\{2, 3, ..., M\}.
	\end{aligned}
	\end{equation}	
	
	Given the fact that all the training samples used by classifier definitely come from generators, meaning $\sum_{m=1}^{M}C^{m}_d(\bx)=1$, we can get the optimal solution of classifier $C_d^m$, that is:
	\begin{equation}\label{eq prop1_}
	\begin{aligned}
	{C^m_{d}}^*(\bx;\theta_d)&=\dfrac{\pi_m P_{G^m_d}(\bx)}{\sum_{i=1}^{M}\pi _iP_{G^i_d}(\bx)}, m \in\{1, 2, ..., M\}.
	\end{aligned}
	\end{equation}
	This concludes the proof.
\end{IEEEproof}

\section{Proof of Proposition 2}\label{apend2}
\begin{IEEEproof}
	Let the optimal discriminator be ${D_d}^*$. Here, we directly use the conclusion in \cite{GAN} that ${D_d}^*=\dfrac{P_{data}^d(\bx)}{P_{data}^d(\bx)+P_{model}^d(\bx)}$.
	Denote the quantity function of generators as $V({\boldsymbol G})$. According to the multi-player zero-sum game in (\ref{eq localMixGAN}), given the optimal discriminator ${D_d}^*$ above and classifier ${C_d}^*$ in proposition 1, generators try to minimize:
	\begin{equation}\label{eq V(G)}
	\begin{aligned}
	V({\boldsymbol G}(\bx;\omega_d))=&\mathbb{E}_{\bx\sim P_{data}^d}[\log\dfrac{P_{data}^d(\bx)}{P_{data}^d(\bx)+P_{model}^d(\bx)}]\\
	&+\mathbb{E}_{\bx\sim P_{model}^d}[\log\dfrac{P_{model}^d(\bx)}{P_{data}^d(\bx)+P_{model}^d(\bx)}]\\
	&-\lambda\{\sum_{m=1}^{M}\pi_m\mathbb{E}_{\bx\sim P_{ G^{m}_d}}[\log\dfrac{\pi _m P_{ G^{m}_d}(\bx)}{\sum_{i=1}^{M}\pi _i P_{ G^{i}_d}(\bx)}]\}.
	\end{aligned}
	\end{equation}
	
	The first two terms in the right-hand of  (\ref{eq V(G)}) is same as in the standard GANs \cite{GAN}, which equals to $2\cdot \mbox{JSD}(P_{data}^d\|P_{model}^d)-\log4$. Here, we focus on the last term of this equation to clarify the effect of the auxiliary classifier.
	
	In light of the conclusion in standard GANs, we can formulate (\ref{eq V(G)}) as follows:
	\begin{equation}\label{eq V2}
	\begin{aligned}
	V({\boldsymbol G}(\bx;\omega_d))=&2\cdot \mbox{JSD}(P_{data}^d\|P_{model}^d)-\log4\\
	&-\lambda\{\sum_{m=1}^{M}\pi_m\mathbb{E}_{\bx\sim P_{ G^{m}_d}}[\log\dfrac{\pi _mP_{ G^{m}_d}(\bx)}{\sum_{i=1}^{M}\pi_i P_{ G^{i}_d}(\bx)}]\}
	\end{aligned}
	\end{equation}
	
	In order to convey the proof more clearly, we use $-\lambda\{\hat V\}$ to denote the last term  in (\ref{eq V2}), and let $\hat{P_G}(\bx)=\sum_{m=1}^{M}\pi _mP_{ G^{m}_d}(\bx)$. Then, We have:
	\begin{subequations}\label{eq Vhat}
		\begin{align}
		\hat V=&\sum_{m=1}^{M}\pi_m\mathbb{E}_{\bx\sim P_{ G^{m}_d}}[\log\dfrac{\pi _mP_{ G^{m}_d}(\bx)}{\sum_{i=1}^{M}\pi _iP_{ G^{i}_d}(\bx)}]\\
		&=\sum_{m=1}^{M}\pi_m\mathbb{E}_{\bx\sim P_{ G^{m}_d}}[\log \dfrac{P_{ G^{m}_d}(\bx)} {\hat{P_G}(\bx)}]
		+\sum_{m=1}^{M}\pi_m\log\pi_m\\
		&=\pi_1\mathbb{E}_{\bx\sim P_{ G^{1}_d}}[\log \dfrac{P_{ G^{1}_d}(\bx)} {\hat{P_G}(\bx)}] + ... \nonumber \\
		&\;\;\; + \pi_M\mathbb{E}_{\bx\sim P_{ G^{M}_d}}[\log \dfrac{P_{ G^{M}_d}(\bx)} {\hat{P_G}(\bx)}]
		+\sum_{m=1}^{M}\pi_m\log\pi_m.
		\end{align}
	\end{subequations}
	
	Using the standard definition of KL divergence, the above formula can be expressed by the sum of multiple KL divergences and the negative entropy of $\bpi$, that is:
	\begin{subequations}\label{eq Vhat2}
		\begin{align}
		\hat V=&~\pi_1\cdot \mbox{KL}(P_{ G^{1}_d}(\bx) \| \hat{P_G}) + ... \nonumber \\
		&\;\;\; + \pi_M\cdot \mbox{KL}(P_{ G^{M}_d}(\bx) \| \hat{P_G})
		+\sum_{m=1}^{M}\pi_m\log\pi_m\\
		&=\sum_{m=1}^{M}\pi_m \mbox{KL}(P_{ G^{m}_d}(\bx) \| \hat{P_G})+\sum_{m=1}^{M}\pi_m\log\pi_m\\
		&=\mbox{JSD}_{\pi_{1},\pi_{2},...,\pi_{M}}(P_{ G^{1}_d},P_{ G^{1}_d},...,P_{ G^{M}_d})+\sum_{m=1}^{M}\pi_m\log\pi_m,
		\end{align}
	\end{subequations}
	where $\mbox{JSD}_{\pi_{1},\pi_{2},...,\pi_{M}}(P_{ G^{1}_d},P_{ G^{1}_d},...,P_{ G^{M}_d})$ is the generalized Jensen-Shannon divergence \cite{generalizedJSD}.
	Therefore, we can rewrite the quantity function of generators in equation (\ref{eq V(G)}) as:
	\begin{subequations}\label{eq VhatJSD}
		\begin{align}
		V({\boldsymbol G}(\bx;\omega_d))=&2\cdot \mbox{JSD}(P_{data}^d\|P_{model}^d)-\log4
		-\lambda \{\hat{V}\}\\
		=&2\cdot \mbox{JSD}(P_{data}^d\|P_{model}^d) \nonumber \\
		&\;\;\; -\lambda \cdot \mbox{JSD}_{\pi_{1},\pi_{2},...,\pi_{M}}(P_{G^{1}_d}, P_{G^{2}_d}, ..., P_{G^{M}_d})\nonumber \\
		&\;\;\; -\lambda\sum_{m=1}^{M}\pi_m\log\pi_m-\log4 .
		\end{align}
	\end{subequations}	
	
	
	Ignoring the last two constant terms, we can finally get the objective function of generators in equation (\ref{eq prop2}). Therefore, optimizing the generators is equivalent to minimizing the JD divergence between $P_{data}^d$ and $P_{model}^d$ while maximizing the differences among those generators.
	This concludes our proof.
\end{IEEEproof}

\section{Proof of Theorem 1}\label{apend3}
\begin{IEEEproof}
	In order to further simplify the objective function of generators in Proposition 2, we firstly recast equation of $\hat{V}$ in (\ref{eq Vhat}):
	\begin{equation}\label{eq prfTh1_1}
	\begin{aligned}
	\hat V=&\sum_{m=1}^{M}\pi_m\mathbb{E}_{\bx\sim P_{ G^{m}_d}}[\log\dfrac{\pi _mP_{ G^{m}_d}(\bx)}{\sum_{i=1}^{M}\pi _iP_{ G^{i}_d}(\bx)}].
	\end{aligned}
	\end{equation}
	
	It can be observed that $\hat V\leq 0$ and the equality occurs only if $\dfrac{\pi _mP_{ G^m_d}(\bx)}{\sum_{i=1}^{M}\pi _iP_{ G^i_d}(\bx)}=1$. Therefore, we have $ P_{ G^m_d}(\bx)\neq 0 ~\rightarrow ~\sum_{i\neq m}\pi _iP_{G^i_d}(\bx)=0~\rightarrow ~\forall i\neq m,~P_{G^i_d}(\bx)=0$.
	In this case, different generators are well-separated without overlapping.
	Further, we can rewrite the objective function of generators in Proposition 2 under $\hat V= 0$: 
	\begin{equation}\label{eq prfTh1_2}
	\begin{aligned}
	{{\boldsymbol G}_d}^*(\bx;\omega_d)=&\mathop{\arg\min}\limits_{{\boldsymbol G}}2\cdot \mbox{JSD}(P_{data}^d\|P_{model}^d).
	\end{aligned}
	\end{equation}
	This is a much simpler formulation and we can get its optimal solution directly, that is: $P_{data}^d=P_{model}^d$. When Assumption 1 holds, which constraints the real data distribution to be a mixture of $M$ components $p_d^m(\bx), m=1, 2, ..., M$, we can easily get the results in (\ref{eq theorem1(b)}) and (\ref{eq theorem1(c)}), that is:
	\begin{subequations}\label{eq theorem1_}
		\begin{align}
		&P_{G_d^m}^*(\bx)=p_d^m(\bx), ~\forall m=1, 2, ..., M, \\ &P_{model}^d(\bx)=\sum_{m=1}^{M}\pi_mP_{G_d^m}^*(\bx)=P_{data}^d(\bx).	
		\end{align}
	\end{subequations}
	
	Up to now, we have proved that at the equilibrium point of the multi-player zero-sum game, generators can synthesize a mixture of traffic types, with the same distirbution as real in a local dataset. In order to evaluate the performance of the classifier, we substitute the well-separated optimal solution $P_{G^m_d}$ into Proposition 1. It can be seen that ${C^m_d}^*$ equals to $1$ when $\bx$ is drawn from $P_{G^m_d}(\bx)$; otherwise, ${C^m_d}^*$ is $0$. Therefore, the optimal classifier can differentiate all the samples produced by different generators correctly. Formally, we have:
	\begin{equation}\label{eq theorem1__}
	\begin{aligned}
	{C_d^{m}}^*(\bx)=
	\begin{cases}
	1, \quad x\sim P_{G_d^w},~w=m\\
	0, \quad x\sim P_{G_d^w},~w\neq m
	\end{cases}	\forall m=1, 2, \ldots, M,		
	\end{aligned}
	\end{equation}
	
	Besides, when $P_{data}^d=P_{model}^d$, meaning generators are synthesizing high-quality samples, the classifier will also be able to classify real samples from the local dataset, e.g., labeling different local services.
	This concludes our proof.
	
\end{IEEEproof}

\end{document}